\documentclass[oneside,11pt,a4paper]{article}
\pdfoutput=1
\usepackage{authblk}
\usepackage[margin=1.2in,a4paper]{geometry}
\usepackage[round,comma]{natbib}
\usepackage{eurosym}

\usepackage[utf8]{inputenc} 
\usepackage[T1]{fontenc}    
\usepackage{amsmath}
\usepackage{hyperref}
\usepackage{cleveref}
\usepackage{nicefrac}       
\usepackage{microtype}      
\usepackage{amssymb}
\newcommand{\R}{\mathbb{R}}

\usepackage{caption}
\usepackage{subcaption}   
\usepackage{todonotes}    
\usepackage{textcomp} 

\newcommand{\fig}[1]{Figure~\protect\ref{#1}}

\renewcommand{\sec}[1]{Section~\protect\ref{#1}}

\newcommand{\bigO}{\mathcal{O}}

\newcommand{\reach}{\textit{reach}}
\newcommand{\grain}{\textit{grain}}
\newcommand{\coherence}{\textit{coherence}}

\title{Comparison-Based Framework for Psychophysics:\\ Lab versus Crowdsourcing}

\author[1]{Siavash Haghiri \thanks{Corresponding author, email: haghiri@informatik.uni-tuebingen.de}}
\author[2]{Patricia Rubisch}
\author[2]{Robert Geirhos}
\author[2]{Felix Wichmann} 
\author[1]{Ulrike von Luxburg}
\date{May 2019}

\affil[1]{Department of Computer Science, University of T{\"u}bingen, Germany}
\affil[2]{Neural Information Processing Group, University of T{\"u}bingen, Germany}

\begin{document}

%\author{ Siavash Haghiri  \and Felix Wichmann \and Ulrike von Luxburg\\
%Department of Computer Science, University of T{\"u}bingen, Germany\\ 
% \{haghiri,ghoshdas,luxburg\}@informatik.uni-tuebingen.de}

\maketitle

\begin{abstract} Traditionally, psychophysical experiments are conducted by repeated measurements on a few well-trained participants under well-controlled conditions, often resulting in, if done properly, high quality data. In recent years, however, crowdsourcing platforms are becoming increasingly popular means of data collection, measuring many participants at the potential cost of obtaining data of worse quality. In this paper we study whether the use of comparison-based (ordinal) data, combined with machine learning algorithms, can boost the reliability of crowdsourcing studies for psychophysics, such that they can achieve performance close to a lab experiment. To this end, we compare three setups: simulations, a psychophysics lab experiment, and the same experiment on Amazon Mechanical Turk. All these experiments are conducted in a comparison-based setting where participants have to answer triplet questions of the form ``is object x closer to y or to z?''. We then use machine learning to solve the triplet prediction problem: given a subset of triplet questions with corresponding answers, we predict the answer to the remaining questions. Considering the limitations and noise on MTurk, we find that the accuracy of triplet prediction is surprisingly close---but not equal---to our lab study.\vspace{1cm}
\end{abstract}

\section{Introduction}

Much of the progress over the last 150 years in experimental psychology and cognitive science stems from well-controlled measurements of human (and animal) behavior. Precise stimulus control has been considered as the \emph{[f]irst commandment of psychophysics} \cite[p.~30]{Geisler_1987}, and the quantitative study of human behavior preferred accuracy over quick (and painless) data acquisition: \emph{The search for short-hand methods in technology is laudable enough, but it is entirely out of place in science, where new trails are being blazed and attempts are made to reduce and to eliminate all of the errors of observation. Ease and convenience are poor experimental guides.}~\citep[p.~656]{Dallenbach_1966}. 

Nonetheless, crowdsourcing experiments are becoming increasingly popular, and there have been positive discussions of using Amazon Mechanical Turk (MTurk) in behavioral experiments \citep[e.g.][]{Chandler_etal_2014,Marder_Fritz_2015}, and suggestions that crowdsourcing may, for some purposes, even be better than traditional laboratory experiments in terms of broader demographics \citep[c.f.][]{Henrich_etal_2010}. However, the worry whether ``behavioral facts'' could ever be obtained without (much) control over hardware, observer concentration, attention, viewing distance, language competence, etc., or even distractions by peers or children and multi-tasking while doing online experiments, are very serious indeed. 

It seems quite clear that a naive approach of simply re-implementing traditional psychophysics experiments on MTurk is likely to fail: the traditional tools such as the just noticeable difference (JND) framework have been developed with the intention to be used in well-controlled lab environments, as they crucially rely on the fine control of the stimuli, and on highly-trained observers. From a purely methodological point of view the reliance on JND-style experiments is reasonable: JND-style data, particularly using the method of forced-choice, are the most reliable and robust estimates of human behavior \citep{Blackwell_1952,Jakel_Wichmann_2006,Wichmann_Jakel_2017}. It would not come as a surprise if such methods failed in crowdsourcing setups. 

For this reason we believe it misleading to ask {\em whether} it is possible to perform psychophysics in a crowdsourcing setup; one would have to ask {\em how the basic methods in psychophysics need to be changed in order to enable crowdsourcing as a new tool for psychophysics}.

In this paper, we suggest an alternative approach to data acquisition in cognitive science that is based on triplet comparisons \citep{Torgerson_1958}. Rather than attempting accurate quantitative measurements of a particular phenomenon, the comparison-based approach we follow aims for qualitative observations of the form ``Stimulus X is more similar to stimulus Y than to stimulus Z''. The obvious potential of such an approach is that the statements may be less dependent on the fine details of the experimental setup, and that the issue of scaling answers across many diverse participants becomes easier. However, there are two major obstacles as well: First, the data acquisition is painful because for a set of $n$ stimuli, there are of the order $n^3$ many triplets. For a data set of size $n=100$ it is impossible to ask participants of a psychological study to answer all triplet comparisons. The second issue is that, so far, a thorough simulation-based evaluation of comparison-based data from a psychophysics perspective has not been performed. 

The approach of triplet comparisons---the method of triads---is not new to psychophysics; in fact, there has been a very long tradition in psychology to explore methods to estimate perceptual (difference) scales from clearly visible \emph{supra-threshold} differences in stimulus appearance \citep[e.g.][]{Torgerson_1958,Coombs_etal_1970,Marks_Gescheider_2002}. Maximum likelihood difference scaling (MLDS), e.g.\ is a popular method in vision science for the estimation of perceptual (difference) scales \citep{Maloney_Yang03,Knoblauch_Maloney_2010}. MLDS can be used together with the method of triads, and there have been reports that both naive as well as seasoned observers find the method of triads with supra-threshold stimuli intuitive and fast, requiring less training \citep{Aguilar_etal_2017}.

In this paper, we run simulations, an experiment in a psychophysics lab and the same experiment on MTurk to address the following main questions: {\em (1) Subsampling approach for triplets:} does there exist a regime such that if we ask participants for $m$ triplets over $n$ objects, we can predict the answers to the remaining triplets with high accuracy? Only if such an approach can be realized in practice, comparison-based methods get into the reach of psychophysics. We use different subsampling strategies and different triplet prediction algorithms and compare them in the three different setups (simulations, lab, MTurk).  {\em (2)  Lab versus crowdsourcing: } We run the same psychophysics task in a well-controlled lab environment (few  participants, each participant measured on a number of different days, and without distractions) and on MTurk (many participants, none of them highly motivated, and each participant only measured once) and compare the results.

The tools we use to address these two questions come from the field of machine learning. Here, algorithms that can deal with comparison-based data have become increasingly popular~\citep{Schultz_Joachims2003,AgaEtAl07,tamuz2011adaptively,Ailon11,Jamieson_Nowak2011,KleLux14,TerLux14,Arias15,JaiJamNow16,haghiri17a}.  

In particular, comparison-based approaches have been used successfully in crowdsourcing \citep{heikinheimo2013crowd,WilKwaBel14,ukkonen2015crowdsourced}.

As our main contribution, we provide empirical results that suggest the feasibility of crowdsourcing with the comparison-based approach for psychophysics experiments as an easier alternative to well-controlled lab experiments. Moreover, we provide guidelines for using comparison-based methods by presenting results about how many triplets are needed, which subsampling approaches lead to better triplet predictions and which methods provide good triplet prediction accuracy. We open doors towards ordinal embedding methods for evaluation of ordinal data in psychophysics experiments. Our case study shows that the embedding can potentially lead to new insights in psychophysics. 
  
\section{Triplet prediction: the key to efficient data collection in psychophysics}

Consider a fixed set $S$ of $n$ items. For $x,y,z \in S$, a triplet question $t(x,y,z)$ is a question of the form "Is $x$ closer to $y$ or to $z$?". The answer to such a question will be denoted by $t(x,y,z)=1$ in the first case and $0$ in the other case. To be able to state the triplet completion problem precisely, we only need to assume that for each triplet $t(x,y,z)$ there exists a probability $p_{xyz} \in [0,1]$ such that $P(t(x,y,z)=1) = p_{xyz}$. In addition, we require $p_{xyz} = 1 - p_{xzy}$ for symmetry reasons. We do not assume a deterministic ground truth, nor the existence of an underlying distance function that is related to the triplet answers, nor a specific noise model. Given a subset $T$ of triplet questions over $S$, the triplet prediction problem is to use the answers to the triplets in $T$ to predict the answers to all remaining triplets with as few errors as possible. 

From a psychophysicist's point of view, the key question is how many triplets we need to ask to a participant, such that the prediction error for the remaining triplets is within an acceptable limit. Naturally, the answer to this question depends on various aspects such as assumption on the data, the distance function (if exists), the noise, the way the triplets in $T$ have been sampled, and the algorithm used to solve the triplet prediction problem. If we do not make any assumptions on the relationship between the triplets, then triplet prediction is impossible. A rather mild assumption is that there exist scores  $\delta(x,y)$ such that $\delta(x,y) < \delta(x,z)$ implies that $P( t(x,y,z)=1) > 0.5$ (that is, the ranking of the scores indicates the most likely answer). In this scenario, it makes sense to use ranking algorithms to solve the triplet prediction problem, which then amounts to ranking the $\binom{n}{2}$ pairwise scores or distances. In the noiseless case and if we can adaptively compare any two distances, standard sorting algorithms require $\bigO(n^2 \log n)$ comparisons to sort $\Theta(n^2)$ distances. If we only get to see noisy triplet comparisons, results by \citet{NegOhSha12} suggest that $\bigO(n^2 \log n)$ comparisons are also sufficient to predict the remaining triplets with small error.

On the other hand, if we make the very strong assumption that the items live in $\R^d$ and triplets are answered according to Euclidean distances with some additional noise, then only $\bigO(d n \log n)$ triplet comparisons are enough~citep{JaiJamNow16}. 

Corresponding to these two extreme cases, we employ two different classes of algorithms to solve the triplet prediction problem: ranking and ordinal embedding algorithms. As ranking methods, we use SerialRank~\citep[SR,][]{FajAspVoj14}, Rank Centrality~\citep[RC,][]{NegOhSha12} and a simple counting method that ranks according to the number of wins in pairwise comparisons, as analyzed by \citet{ShaWai15}. For embedding, we use Generalized Non-Metric Multidimensional Scaling \citep[GNMDS,][]{AgaEtAl07}, (t-)Stochastic Triplet Embedding \citep[STE/t-STE,][]{MaaWei12} and Soft Ordinal Embedding \citep[SOE,][]{TerLux14}.

In our experiments, we use three different {subsampling strategies} to generate a subset $T$ of triplets to be used as input to the triplet prediction problem. (1) \textit{Random:} We sample $m$ triplets uniformly at random from the set of all possible triplets. (2) \textit{$l$-repeated-random:} We sample $m/l$ triplets uniformly at random from the set of all possible triplets, and ask each triplet $l$ times. (3) \textit{Landmark:} We fix a small number of landmark objects $\ell_1, ..., \ell_k$, then ask all triplet questions of the form $(x, \ell_i, \ell_j)$.

\section{Triplet prediction problem: Simulations}
\label{sec:simlulations}

In this section, we use simulations to evaluate the feasibility of triplet prediction on toy data sets. We use two different data sets: points sampled uniformly at random from the $d$-dimensional unit cube, and MNIST handwritten digits. In both cases, we use Euclidean distances as ground truth for answering triplet questions. The setup of our simulations is to first sample a set of $n$ objects, then sample a set of $m$ triplets over those objects (with different sampling strategies), and solve the triplet prediction problem by various algorithms. In the end, we measure the accuracy of triplet prediction on a new subsample of triplets. In some simulations we additionally perturb the correct triplet answers (as given by the Euclidean distance) by noise: for a fixed noise probability $p < 0.5$, we flip the answers to triplet questions with probability $p$, for each triplet independently.

\subsection{Triplet prediction methods}
For the triplet prediction task, we use the embedding and ranking methods mentioned above. First, we evaluate how all these triplet prediction methods perform for different numbers $n$ of items. In this simulation, the number of triplets grows like $\Theta(n \log n)$ (this is the regime in which ordinal embedding methods are supposed to have constant error). We report accuracy as the fraction of all possible triplets that are correctly predicted. The data sets we use are randomly generated points in the $3$-dimensional unit cube and randomly selected subsets of MNIST. We report the results for MNIST in the supplementary material. Unless noted otherwise, triplet answers are generated according to Euclidean distance and embeddings are in $3$ dimensions.

\begin{figure}
\newcommand{\myimagewidth}{1.2cm}
    \centering
    \begin{subfigure}{0.55\textwidth}
    \centering
    \includegraphics[width=\linewidth]{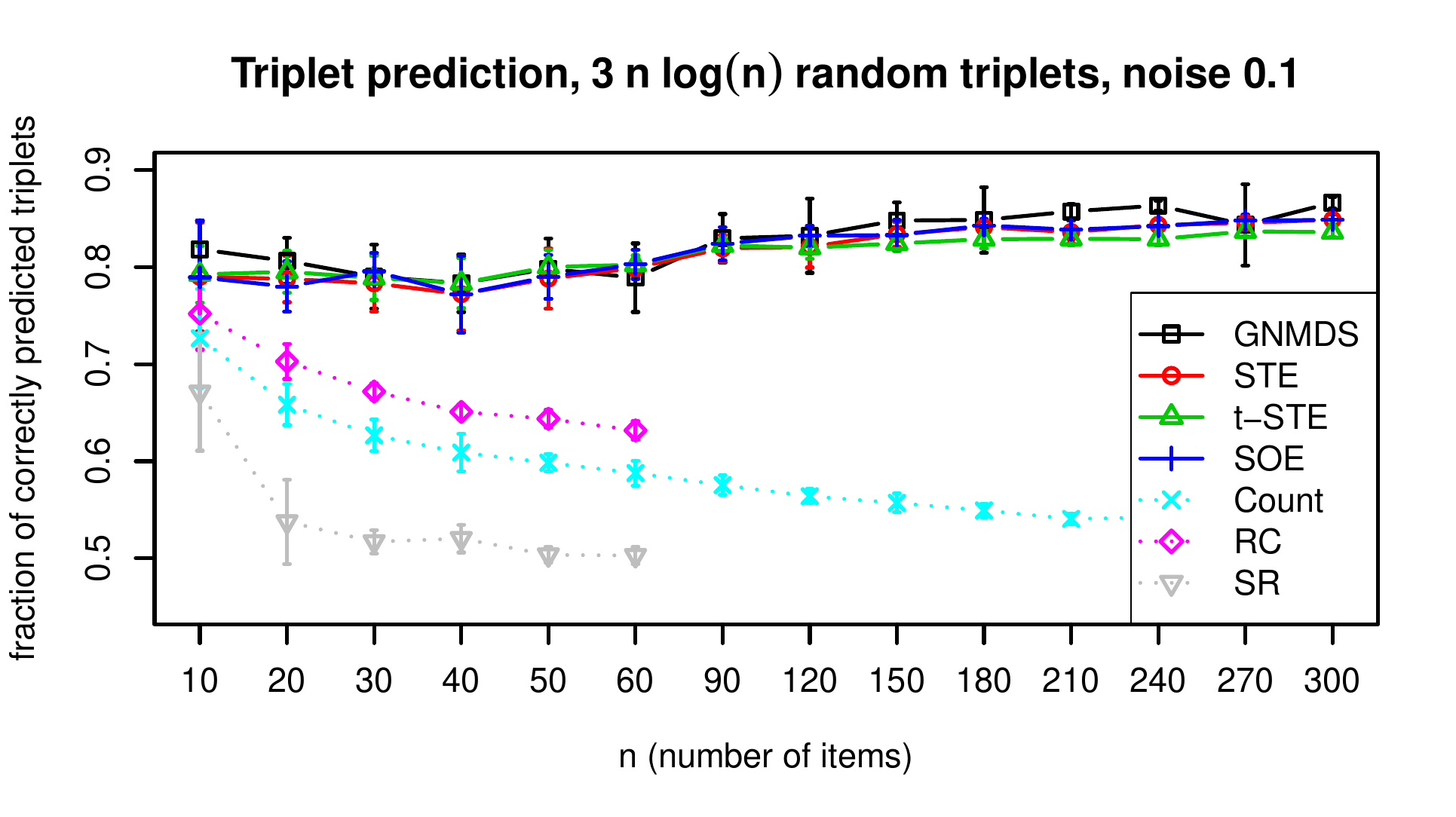}
    \end{subfigure}
    \begin{subfigure}{0.14\textwidth}
    \centering
    \includegraphics[width=\myimagewidth]{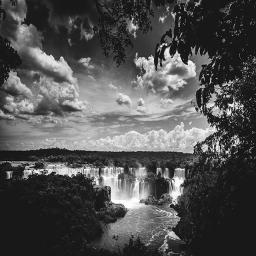}
    \end{subfigure}
    \begin{subfigure}{0.29\textwidth}
    \centering
    \includegraphics[width=\myimagewidth]{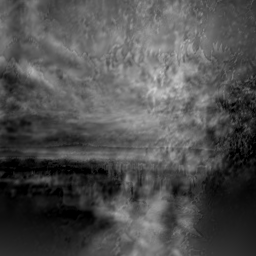} \\[0.1cm]
    \includegraphics[width=\myimagewidth]{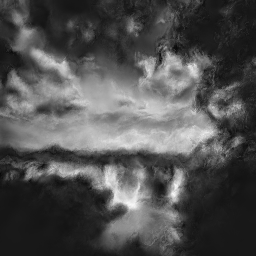} \hspace{0.0cm}
    \includegraphics[width=\myimagewidth]{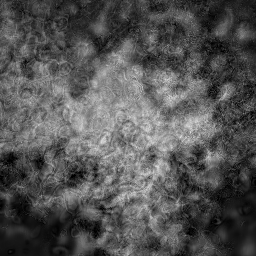}
    \end{subfigure}
    
    \vspace{-2mm}    
    \caption{Left: Triplet prediction accuracy for various embedding and ranking methods. The number of triplets shown to the algorithm depends on the number of items $n$, which varies along the horizontal axis. Items are generated as points uniformly at random in $[0,1]^3$. We show the averages over $10$ runs. Middle: The original image from our eidolon experiment (see \sec{sec:eidolon_experiment}). Right: An example triplet question --- \textit{``Which of the bottom two images is more similar to the top image?''}}
    \label{fig:compare_methods_synthetic_eidolon_example}
\end{figure}

In \fig{fig:compare_methods_synthetic_eidolon_example} (left), we observe that the prediction capability of embedding algorithms is about constant if $3 n \log_2(n)$ triplets are known. This demonstrates that the error bound of \citet{JaiJamNow16} also seems to hold for embedding algorithms that do not perform empirical risk minimization. We verify that the accuracy does not decrease for up to 300 items. In our later experiments, we will have 100 items and 2000 triplets. %
Since $3 \cdot 100 \log_2(100) \approx 2000$, we can expect a meaningful embedding from 2000 triplets. Moreover, the different ordinal embedding algorithms often perform quite similar to each other, so in the following we only report the result of one of them, SOE. 
We also see that ranking algorithms fail to achieve good prediction accuracy with so few triplets, which is not surprising as they can not exploit the geometry of the data and need to rank $\Theta(n^2)$ distances. However, ranking algorithms do perform well if $\Theta(n^2 \log n)$ triplets are known (see additional results in the supplementary material, where we also report similar behavior on MNIST data). For many applications, including our experiments below, it is not feasible to collect such a large number of triplets, though. %

\subsection{Subsampling strategies}

\paragraph{Random vs $l$-repeated-random} We vary the number of items $n$, the number $l$ of repetitions per triplet and the noise level $p$, %
i.e., the probability that a training triplet is incorrect. For the $l$-repeated-random strategy, we use $2000$ triplets, each of which gets answered $l$ times. For the random strategy, we use $2000 \cdot l$ triplets that are randomly sampled from all triplets (sampling is with repetitions, but we do not repeat triplets systematically as in the $l$-repeated-random setting). Here, we show the results of the SOE algorithm for triplet prediction in order to keep the plots clear. Answers to triplets that are systematically repeated $l$ times are aggregated via majority vote, i.e., the answer to a triplet that we provide as input to SOE is the same that is given in more than $\frac{l}{2}$ of the $l$ repetitions of that same triplet (note that we always choose $l$ to be odd).%

\begin{figure}
\centering
\begin{subfigure}{0.29\textwidth}
    \centering
    \includegraphics[height=4.0cm]{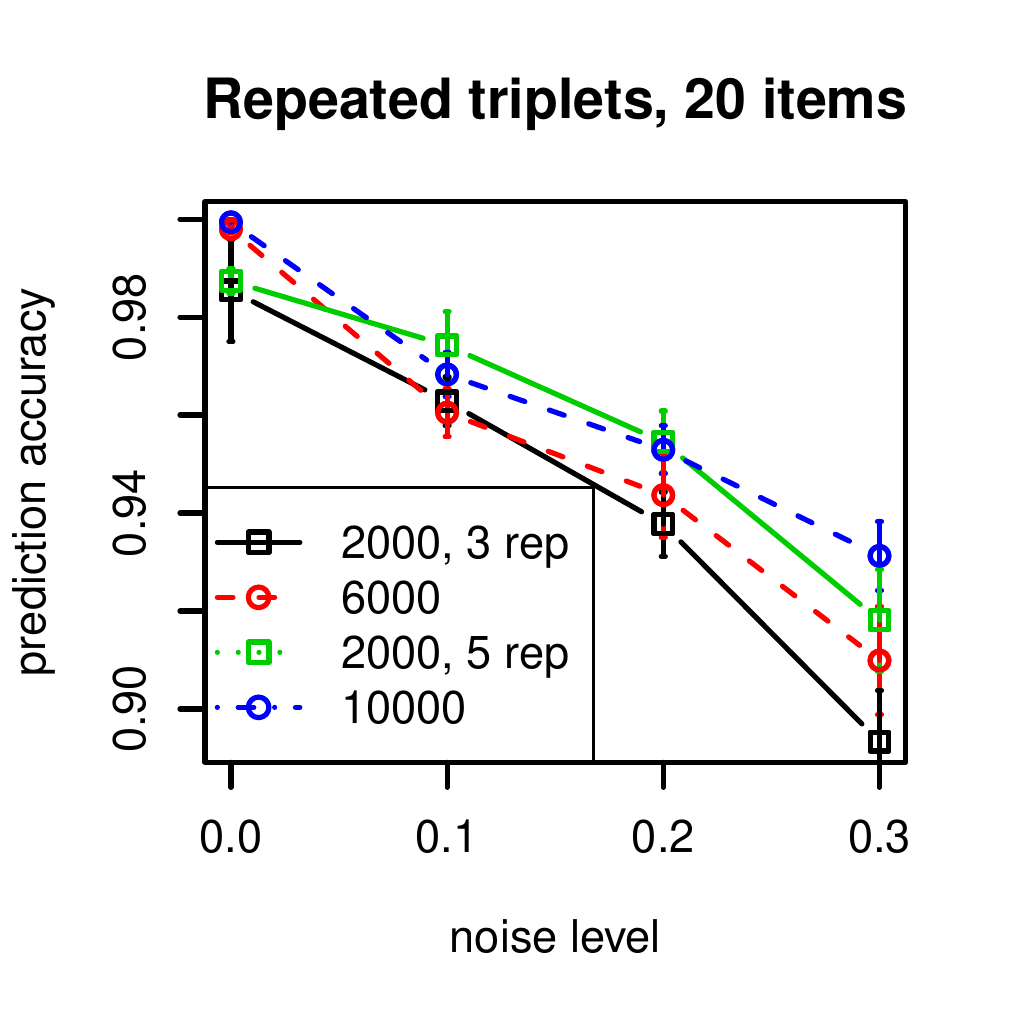}
\end{subfigure}
\begin{subfigure}{0.29\textwidth}
    \centering
    \includegraphics[height=4.0cm]{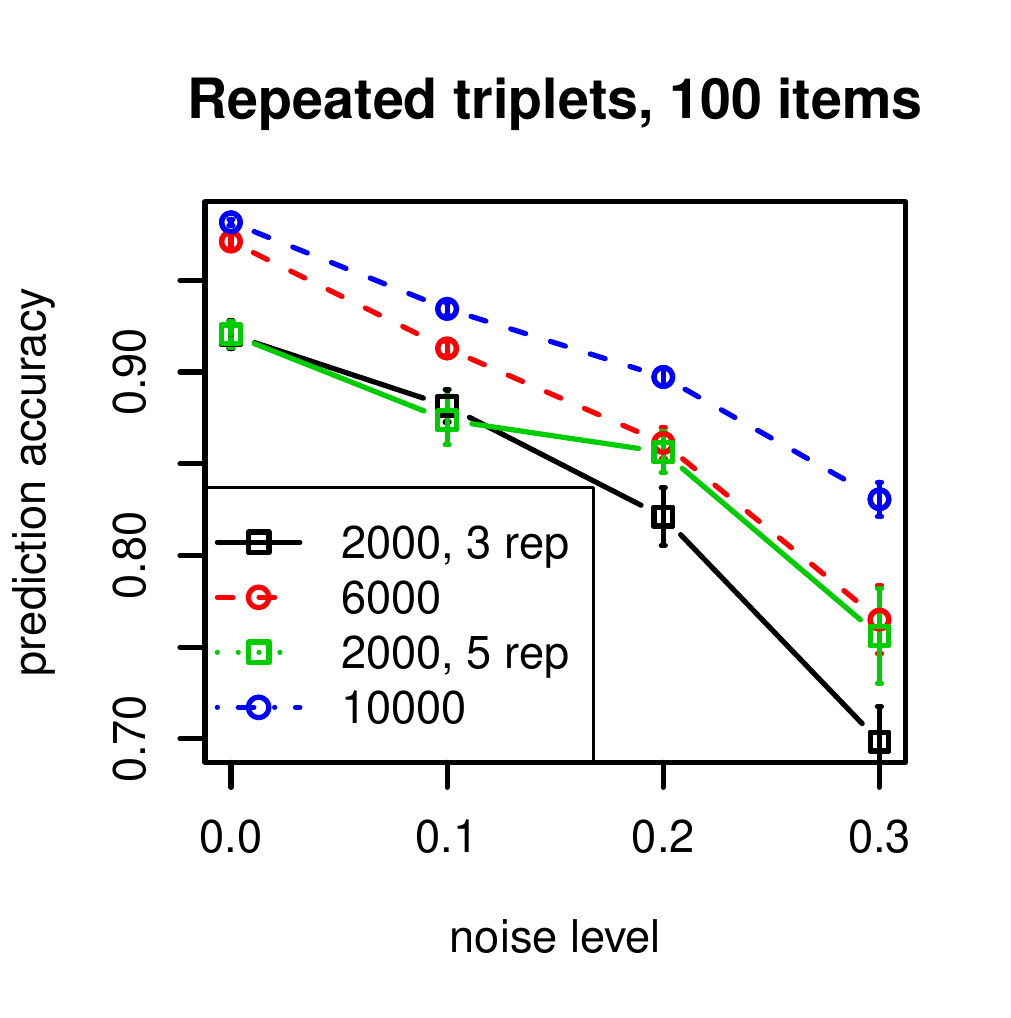}
\end{subfigure}
\begin{subfigure}{0.4\textwidth}
    \centering
    \includegraphics[height=4.0cm]{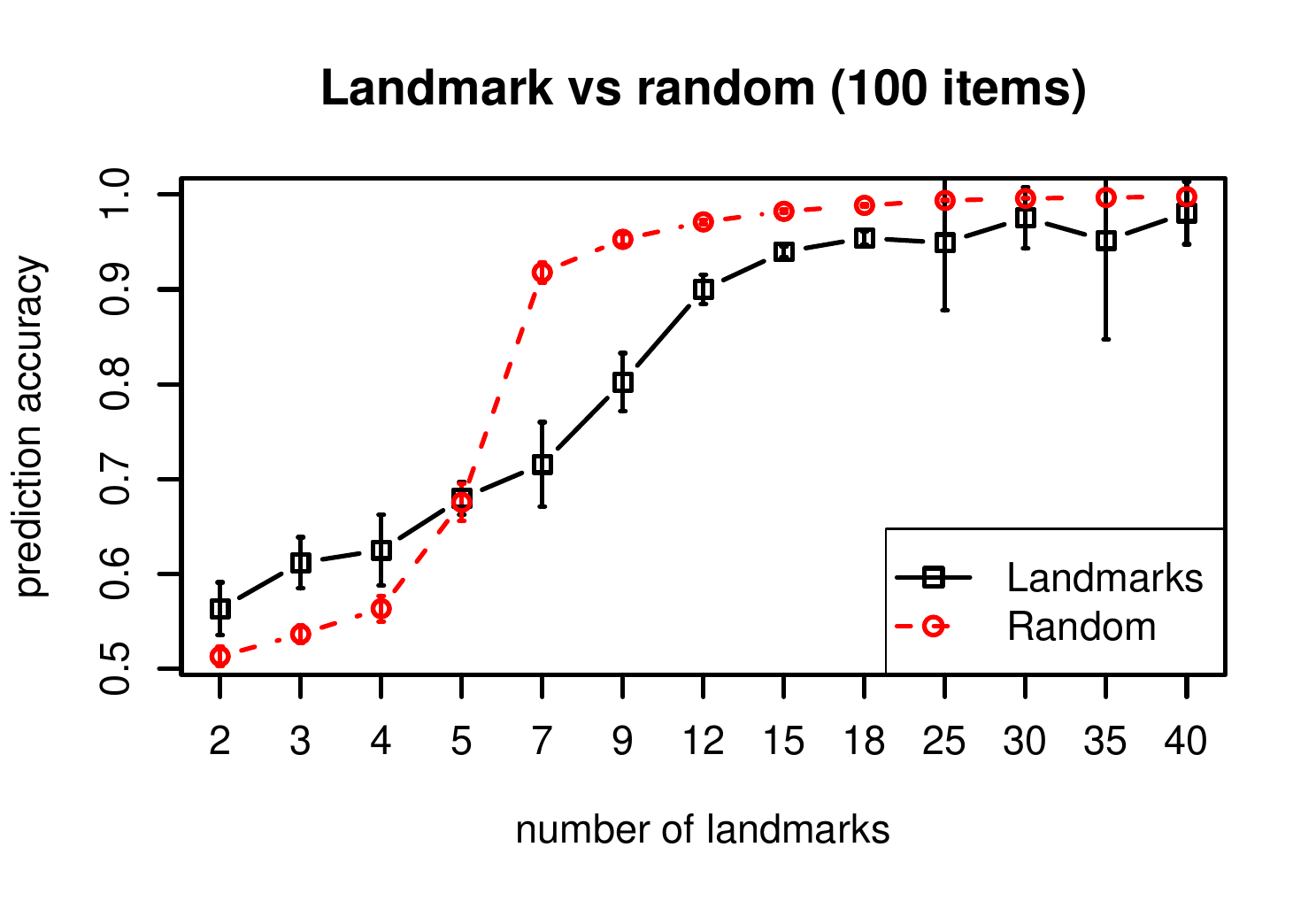}
    \end{subfigure}
\vspace{-2mm}
\caption{Left, middle: The vertical axis shows the accuracy in predicting all triplets from a subset of triplets using SOE embedding. 20 resp. 100 items are sampled uniformly at random in $[0,1]^3$. A subset of all possible triplets is selected for embedding. Either 2000 triplets are answered $l \in \{3,5\}$ times repeatedly (black and green) or $l \cdot 2000$ triplets are answered once (red and blue)%
, noise is displayed on the horizontal axis. For repeated triplets, answers are aggregated using majority vote. Right: Triplet prediction accuracy for the landmark setting versus the random setting. %
}
\label{fig:rep_vs_random_by_noise}
\end{figure}

In \fig{fig:rep_vs_random_by_noise}, we show the results of triplet prediction with SOE embedding and $l$-repeated-random sampling, varying $l$, $n$ and the noise level $p$. In the left subfigure, we see that repeated sampling and employing majority vote may slightly improve prediction accuracy if the noise level is moderate ($0.1$ or $0.2$) and the number of triplets is relatively high in relation to the number of items. For a higher noise level, random sampling outperforms repeated sampling, which is in accordance with results by \citet{LinMauWel14} that repeated answers with majority voting may be beneficial especially for moderate noise levels. In a situation that we consider more interesting, namely if the number of given triplets is low relative to the number of items (middle subfigure), repeated triplets lead to worse prediction accuracy than the same number of distinct triplets.

\paragraph{Random vs landmark} We select $k$ landmarks $\ell_1, \dots, \ell_k$ uniformly at random from all $n = 100$ items and evaluate $t(x, \ell_i, \ell_j)$ for all $1 \leq i < j \leq k$, $x \in S \setminus \{\ell_i, \ell_j\}$ to produce input triplets for embedding. We use the same number of triplets, i.e. $\binom{k}{2}(n-2)$, chosen uniformly at random to construct another embedding for comparison. Similar to the repeated strategy, we only show SOE results in \fig{fig:rep_vs_random_by_noise} (right) for clarity.

The results show that using the landmark strategy for triplet subsampling leads to a worse triplet prediction accuracy than random triplet sampling, except in a regime when very few triplets are known and the predictions are hardly better than random guessing.

\section{The eidolon experiment: Lab and MTurk}
\label{sec:eidolon_experiment}

As a representative psychophysics task we choose a comparison task between images. To generate the images, we use the Eidolon Factory by \citet{KoeEtAl17}---more specifically, its \mbox{partially\_coherent\_disarray()} function. In this toolbox, a given basis image can be  distorted systematically, using three different parameters called \reach, \grain{} and \coherence. An eidolon of a basis image then corresponds to a parametrically altered version of this image. Reach controls the strength of a distortion (the higher the value, the stronger the amplification), grain modifies how fine-grained the distortion is (low values correspond to `highly fine-grained'), whereas a parameter value close to 1.0 for coherence indicates that ``local image structure [is retained] even when the global image structure is destroyed'' \citep[p.~10]{KoeEtAl17}. From a psychophysics point of view, we want to know which and to what degree the image modifications influence the percept. Starting with a black and white image of a natural landscape as basis image (see \fig{fig:compare_methods_synthetic_eidolon_example}, middle), we generate 100 altered images, using reach and grain in $\{5, 12, 26, 61, 128\}$ and coherence in $\{0.0, 0.33, 0.67, 1.0\}$.%

\paragraph{Lab experiment setup} In our psychophysical lab, we ask eight participants (4 male, 4 female, aged 19 to 25, mean 21 years) to answer triplet questions. See \fig{fig:compare_methods_synthetic_eidolon_example} (right) for an example question. For this purpose, participants use a standard computer mouse to click on the one of the two bottom images that they deem more similar to the top image. Stimuli are presented on a $1920 \times 1200$ pixels ($484 \times 302$ mm) VIEWPixx LCD monitor (VPixx Technologies, Saint-Bruno, Canada) at a refresh rate of 120~Hz in an otherwise dark chamber. Viewing distance is 100 cm, corresponding to $3.66 \times 3.66$ degrees of visual angle for a single $256 \times 256$ pixels image. The surround of the screen is set to a grey value of $0.32$ in the $[0, 1]$ range, the mean value of all experimental images. The experiment is programmed in MATLAB (Release 2016a, The MathWorks, Inc., Natick, Massachusetts, United States) using the Psychophysics Toolbox extensions version 3.0.12 \citep{Brainard97, KleEtAl07} along with the iShow library (\url{http://dx.doi.org/10.5281/zenodo.34217}).

Answers have to be given within $4.5$ seconds after a triplet presentation onset, otherwise the triplet will be registered as unanswered and the experiment will proceed to the next triplet. (this occurred in $0.013$\% of all cases only). The experiment is self-paced, i.e., once a participant answers a question, the next one will appear directly after a short fixation time of $0.3$ seconds, during which only a white $20 \times 20$ pixels fixation rectangle at the center of the screen is shown. Before the experiment starts, all test subjects are given instructions by a lab assistant and practice on 100 triplets to gain familiarity with the task. The set of practice triplets is disjoint from the set of experimental triplets. Participants are free to take a break every 200 triplet questions. They give their written consent prior to the experiment and are either compensated \euro 10 per hour for their time or gain course credit towards their degree. All test subjects were students and reported normal or corrected-to-normal vision.

We implement three triplet subsampling strategies: random, 3-repeated-random and landmark. Every participant answers 6000 triplet questions in three approximately 75-min sessions of 2000 triplets each. In the landmark subsampling experiment we have a fourth session during which participants answer 1500 triplet questions as a test set. Subjects 1, 2 and 3 each answer 6000 triplets, which are sampled uniformly at random (with repetitions). Subjects 4, 5 and 6 each answer 2000 triplets which are repeated 3 times. The triplets for the first sessions of subjects 1 and 4 are identical and presented in the same order. In their second and third session, subject 4 answers the same triplets as in their first session, but the order is randomized in each session. In the same way, subject 5 answers a subset of subject 2's triplets and subject 6 answers a subset of subject 3's triplets.

The triplets for subjects 7 and 8 are selected according to the landmark strategy. We select 12 landmarks, giving rise to $\binom{12}{2}(100-2) = 6468$ triplets, from which we subsample 6000 triplets that are presented to the subjects over three sessions. The landmarks are picked by hand, such that they provide a good covering of the parameter space. The triplets were identical for both subjects. Subjects 7 and 8 also answer 1500 triplets sampled uniformly at random, as a test set.%

\paragraph{MTurk setup}
In order to achieve a fair comparison between the lab and crowdsourcing settings, we design our crowdsourcing experiment on MTurk to be as similar as possible to the lab experiment. To this end, we create a survey page on our local webservers to which all participants get redirected. The page layout, background color and stimuli presentation times are chosen as in the lab experiment.
As in the lab, we generate sessions consisting of 2000 (2500 for the landmark setting) triplet questions.  
The allocated time for sessions is 4 hours, however a session is done in less than 2 hours on average.

To obtain data sets of 6000 triplets, one needs three consecutive sessions from each participant, taking more than 6 hours. In the lab, these sessions were held on separate days. Conducting many repeated sessions with the same participant and precise time schedules is the strength of the lab, which we sacrifice by using the crowd on MTurk. On MTurk, we do not conduct long tasks of 6000 triplets.

To filter out participants that may just give random answers to our triplet questions, we implement a sanity check for validation. We pick 20 triplet questions with very obvious correct answers in every session and mark them as our \textit{Gold Standard Questions} (see supplementary material for an example set of 10 gold standard questions). We excluded participants from the evaluation if their error rate on those questions was larger than 20\%. This was the case for 7 out of 60 sessions (12\%); thus it is as well to note that in a crowdsourcing setting the exclusion of---for whatever reason---poorly performing participants is an issue that typically does not, or to a much lesser extent, affect lab experiments. This is a potentially thorny issue, as one should not, of course, exclude participants because their behavior does not align well with one's hypothesis. Sanity checks for crowdsourcing experiments certainly deserve more thought in the future.%

We run three sets of experiments, corresponding to our three strategies of triplet subsampling. In the first set of experiments, we ask 2000 uniformly random triplets from 30 participants. Out of those, 23 participants passed the sanity check in the end. In the second set of experiments we implement the 3-repeated-random subsampling strategy in 15 sessions each consisting of 2000 triplets. We fix 5 sets of 2000 random triplets and ask each of them three times. All these participants were accepted in the end. The third set of experiments is the landmark setting. Here we use exactly the same triplets as the lab experiments. We divide the 6000 landmark triplets and 1500 test triplets each into 3 sets. In this way, each participant has to perform a session consisting of 2000 landmark triplets and 500 test triplets. We have 15 sessions for this task, and all of them were validated. 

\subsection{Results}

For the eidolon experiment, we do not have ground truth information about the distances. Therefore, we can only measure how well we can predict answers to a set of test triplets using a set of training triplets, in the knowledge that the answers to the test triplets may be noisy.
The 3-repeated-random setting provides some information about that noise level: For lab subjects 4, 5 and 6, for each triplet we can take the majority vote of the answers for that triplet. The agreements between triplet answers and majority votes over these subjects are $90.8\%$, $90.8\%$ and $89.0\%$, respectively. As a result, the highest accuracy that we can expect for predicting participants' triplet answers is about $90\%$.

We have made two observations that apply to all of the following results. First, the t-STE method consistently outperforms other methods for triplet prediction. Secondly, perhaps surprisingly, prediction accuracy is best for embedding in 2 dimensions, even though images are generated using 3 parameters (see supplementary material for detailed results). This may be due to interaction of the parameters. Therefore, we only report results for t-STE and embedding in 2 dimensions. In the following we briefly describe the details of subsampling settings.

\begin{figure}
\newcommand{\myheight}{4.5cm}
\centering
    \begin{subfigure}{0.27\textwidth}
    \centering
    \includegraphics[height=\myheight]{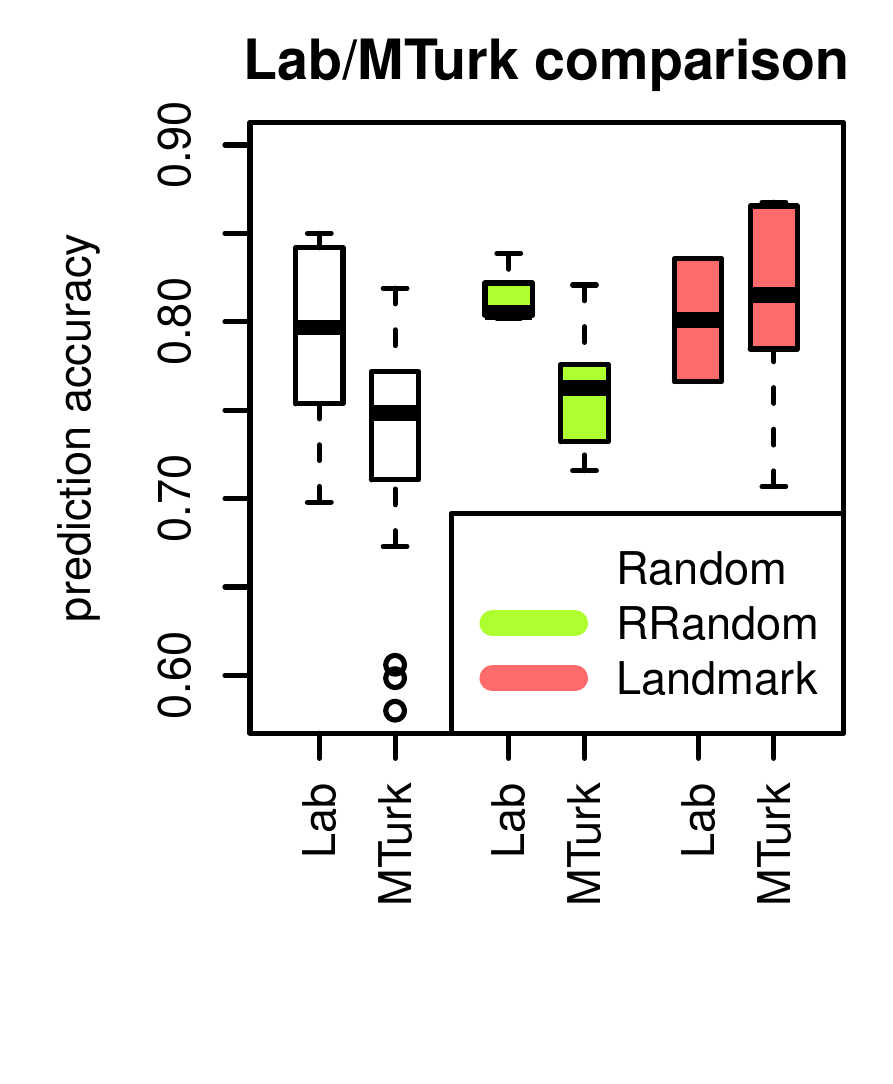}
    \caption{}
    \end{subfigure}
    \centering
    \begin{subfigure}{0.24\textwidth}
    \centering
    \includegraphics[height=\myheight]{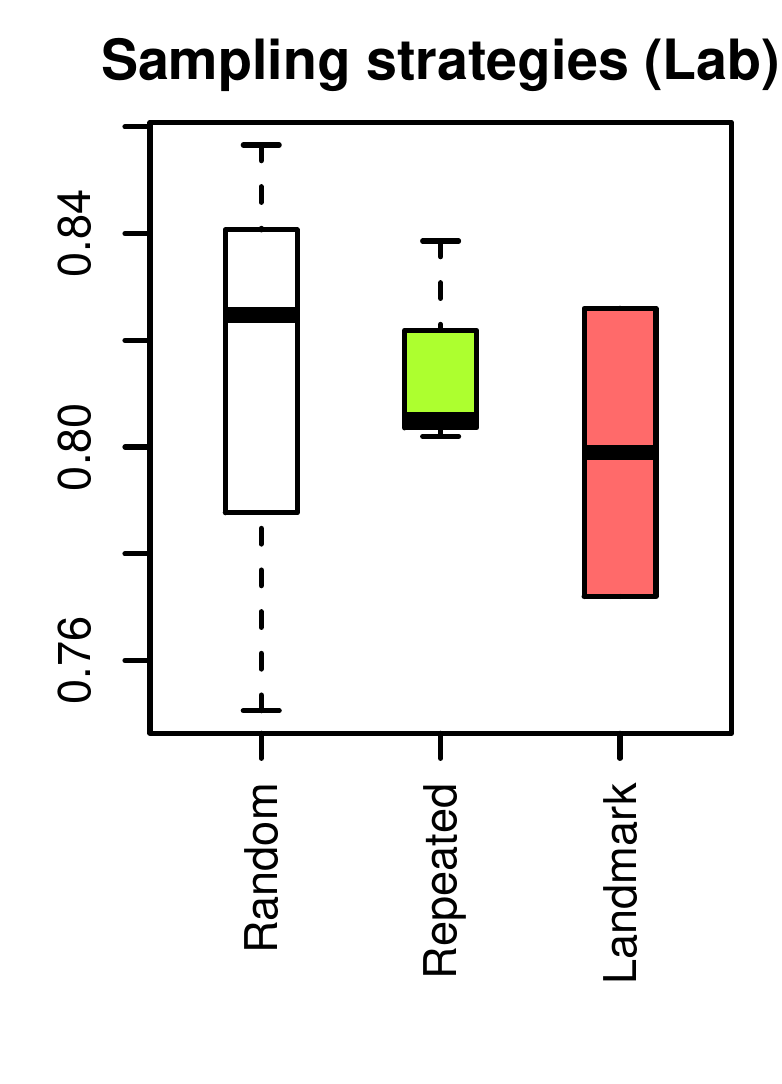}
    \caption{}
    \end{subfigure}
    \begin{subfigure}{0.23\textwidth}
    \centering
    \includegraphics[height=\myheight]{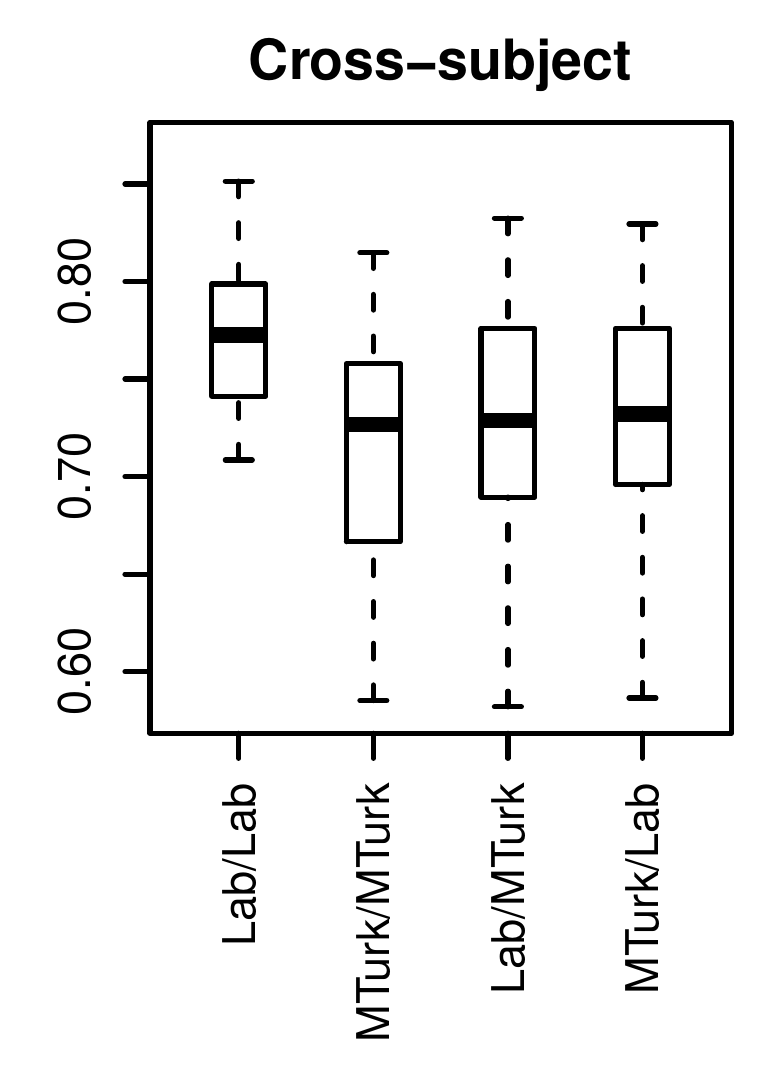}
    \caption{}
    \end{subfigure}
    \begin{subfigure}{0.23\textwidth}
    \centering
    \includegraphics[height=\myheight]{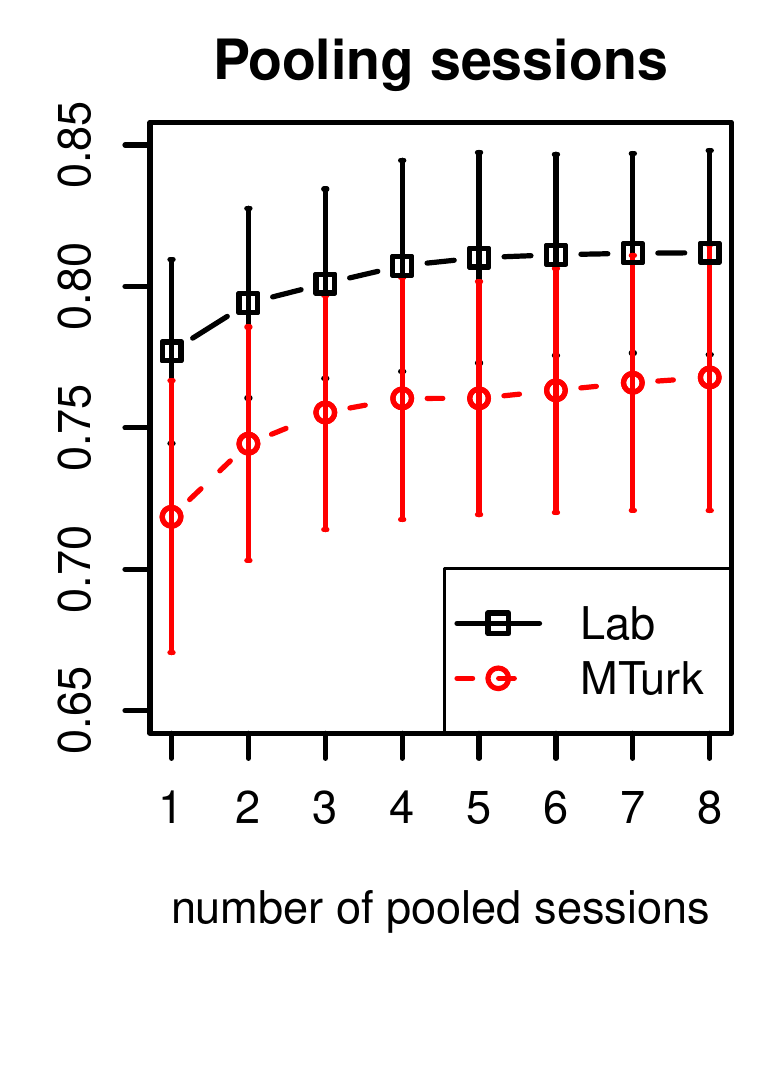}
    \caption{}
    \end{subfigure}
    \centering
    \caption{(a) Triplet prediction accuracy for 3 subsampling strategies. Note that boxes with different colors are not comparable since they have different training sizes. (b) Prediction accuracies for 3 various subsampling strategies from the lab experiment in a comparable setting. (c) Cross-subject prediction accuracies. (d) Prediction accuracy with respect to pooling size.}
    \label{fig:repeated_lab_mturk}
\end{figure}

\textbf{Random:} The two left boxes in \fig{fig:repeated_lab_mturk} (a) show the triplet prediction accuracy for the lab and MTurk experiments. We use 1500 random triplets of each participant for the embedding and 250 triplets for evaluation. For the MTurk experiments we report the performance based on 23 participants who passed the sanity check. \textbf{Repeated-Random}: For the training set we merge 1500 triplets from each session which we ask 3 times. In this way the training set contains 4500 triplets. As the test set we choose 250 random triplets from the remaining triplets of the 3 repeated sessions. The two middle boxes in \fig{fig:repeated_lab_mturk} (a) show the triplet prediction accuracy for both MTurk and lab experiments in this setting. \textbf{Landmark:} The training set contains 6000 triplets of 3 sessions. The test set contains 1500 triplets. The two right boxes in \fig{fig:repeated_lab_mturk} (a) show the triplet prediction accuracy in this setting.

\fig{fig:repeated_lab_mturk} (a) shows that in most cases the MTurk participants' responses can be predicted with about 5 percent less accuracy than those of the lab participants. %
Furthermore, MTurk data have a higher variance. Detailed results for each participant are reported in the supplementary material. Clearly, lab data are of higher quality than MTurk data; however, the difference is certainly not dramatic, and one can imagine scenarios where the ease of data acquisition and the much larger sample of participants outweights the (slightly) lower data fidelity.%

We also report the prediction accuracy of various subsampling strategies for the lab experiments in a comparable setting in \fig{fig:repeated_lab_mturk} (b). We use $4500$ triplets for training, selected according to the respective strategy, i.e., random or $1500$ triplets three times each, or 4500 triplets from the set of landmark triplets, and test prediction accuracy on a test set of 250 random triplets.%

\subsection{Cross-subject analysis}
\label{sec:cross_subject_analysis}
Assuming a ground truth embedding over the eidolon images, we investigate the consistency of the triplets of one participant with the embedding gathered from another participant. We focus on the sessions with random triplets. We use 9 sessions from the first three subjects in the lab experiments and 23 validated sessions from the random strategy in the MTurk experiments. We perform this analysis in 4 parts: Lab/Lab, Lab/MTurk, MTurk/Lab and MTurk/MTurk. Lab/MTurk means that all combinations of sessions from MTurk and sessions from the lab are used for training and testing, respectively. %
Box plots in \fig{fig:repeated_lab_mturk} (c) show the prediction accuracies in four parts of the experiments.

The average accuracy in the first two parts (Lab/Lab and MTurk/MTurk) is only slightly worse than the experiments on the same subject (\fig{fig:repeated_lab_mturk} (c) two left boxes). This strongly suggests that participants---both in the Lab and on MTurk---perceive the similarities between images in a consistent manner. The prediction accuracies convinced us that visual perception in our experimental scenario can be captured in a Euclidean space with two dimensions. Considering this embedding space participants show very similar interpretation of the similarities. The difference between Lab and MTurk is likely due to the noise and uncontrolled setting of data gathering in MTurk. Again the MTurk experiments show a higher variance---this, clearly, is a drawback of crowdsourcing in comparison with lab experiments: Whilst the average difference in prediction accuracy is only around 5\%, the difference is larger than 10\% for the most difficult to predict observers.

\subsection{Pooling triplets} 

In this section, we examine if pooling many triplets can improve the triplet prediction accuracy. We only include triplets from the random strategy, namely 9 sessions from subjects 1 to 3 in the lab and 23 sessions from the MTurk participants who pass the sanity check. We run this experiment on 20 trials. In each trial, we permute the list of sessions in both lab and MTurk session lists. The first session of the permuted list is then used as the test set and the remaining ones are added to the pool of training set one by one. We report the average and standard deviation of prediction accuracy over 20 trials in \fig{fig:repeated_lab_mturk} (d) with respect to the size of training pool. We see a similar pattern in pooling of the sessions for both lab and MTurk experiments. We can roughly gain about 5 percent in accuracy with 4 sessions. However, adding more sessions can not help to improve. We again see about 5 percent difference in the accuracy of MTurk pooled sessions in comparison to the lab pooled sessions.

\section{Ordinal embedding as a tool to evaluate comparison-based data}
\begin{figure}
\centering
    \begin{subfigure}{0.32\textwidth}
    \centering
    \includegraphics[width=\linewidth]{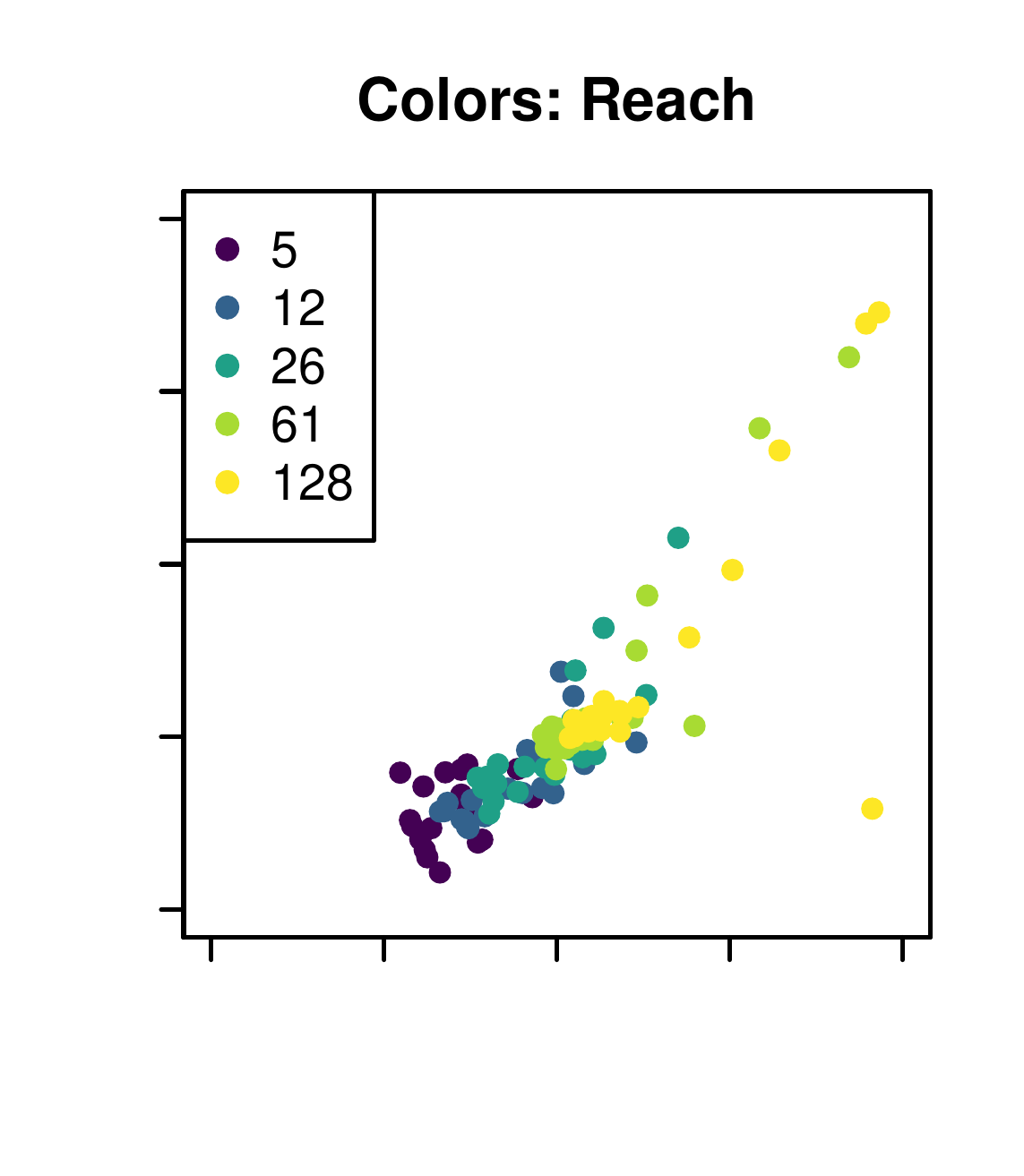}
    \end{subfigure}
    \centering
    \begin{subfigure}{0.32\textwidth}
    \centering
    \includegraphics[width=\linewidth]{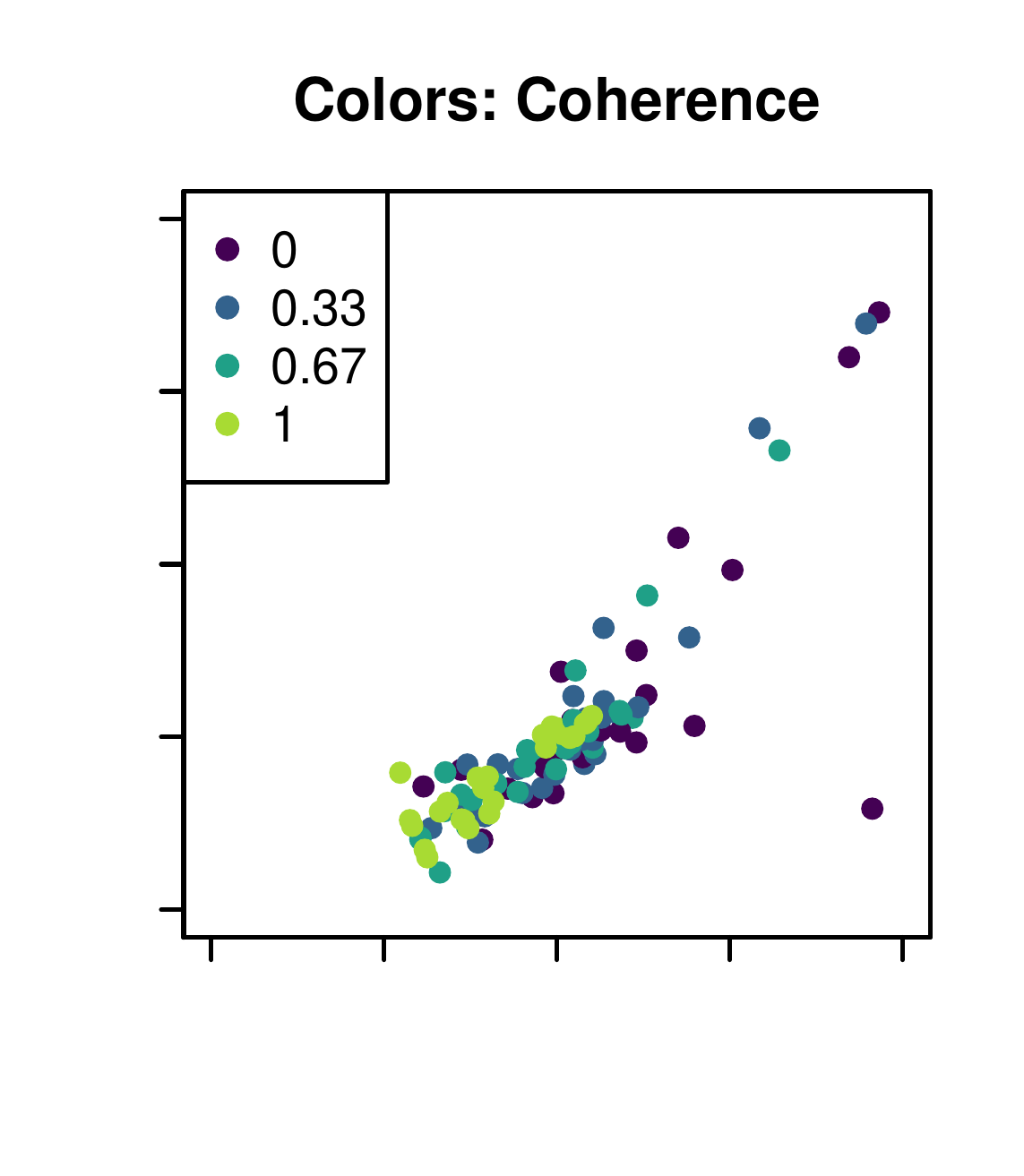}
    \end{subfigure}
    \begin{subfigure}{0.32\textwidth}
    \centering
    \includegraphics[width=\linewidth]{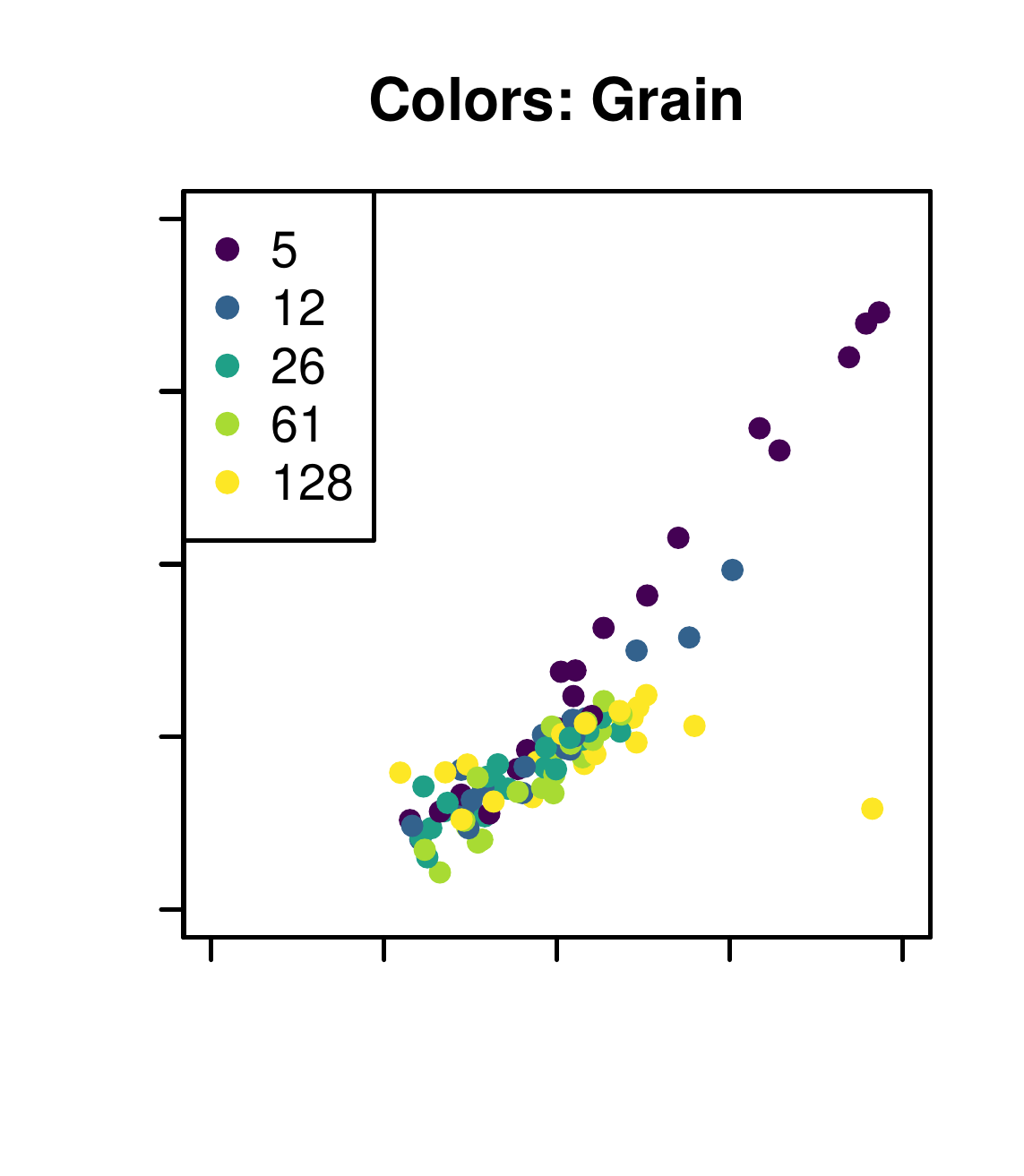}
    \end{subfigure}
    \centering
    \vspace{-10mm}
    \caption{t-STE embedding in $2$ dimensions from all triplets by the first three lab subjects (i.e., all sessions with completely random triplets). Each dot corresponds to one of the 100 images from our eidolon experiment, dots are colored by the values of the three parameters reach, coherence and grain used to generate the images.}
    \label{fig:parameters_in_embedding}
\end{figure}

In the previous sections we have seen that by combining a triplet subsampling strategy with triplet prediction, the data acquisition for comparison-based settings becomes feasible. The key is of course now how we exploit the answers to all triplet questions in psychophysical studies. One straightforward approach is to use Euclidean embeddings to have new insights about psychophysical problems.

While the problem of ordinal embedding was already considered in the 1960s
(e.g., \citealp{Kruskal64}), we want to emphasize the relevance of recent progress in the field of comparison-based data for psychophysical experimenters. Promising new methods yield better results faster and require less data, allowing for experiments on larger scales without compromising trustworthiness of the outcomes.

As a proof of concept, consider ordinal embeddings based on the lab experiment, as shown in \fig{fig:parameters_in_embedding} (also see the appendix for further figures). The three figures show the same embedding (based on the same triplet answers), each point corresponding to one image. The three embeddings are colored in three different ways according to the true parameters of reach, coherence and grain of the object. 
 We can see that one dimension of the embedding space appears to be largely determined by the reach parameter, and to a certain extent by the coherence parameter as well. The second dimension on the other hand seems to be linked to the grain parameter. It will be interesting to see more profound results driven by ordinal embeddings of the images we used in our experiment. However, a thorough analysis and interpretation of the embedding results is outside the scope of this work. 

\section*{Acknowledgements}

The authors thank Guillermo Aguilar for fruitful discussions. This work has been supported by the German Research Foundation DFG (SFB 936/ Z3), the Institutional Strategy of the University of T{\"u}bingen (DFG ZUK 63), and the International Max Planck Research School for Intelligent Systems (IMPRS-IS).

\newpage
\appendix

\section{Additional experimental results}
\subsection{Simulations}
\label{app:simulations}

\begin{figure}
    \centering
    \begin{subfigure}{0.32\textwidth}
    \centering
    \includegraphics[width=\linewidth]{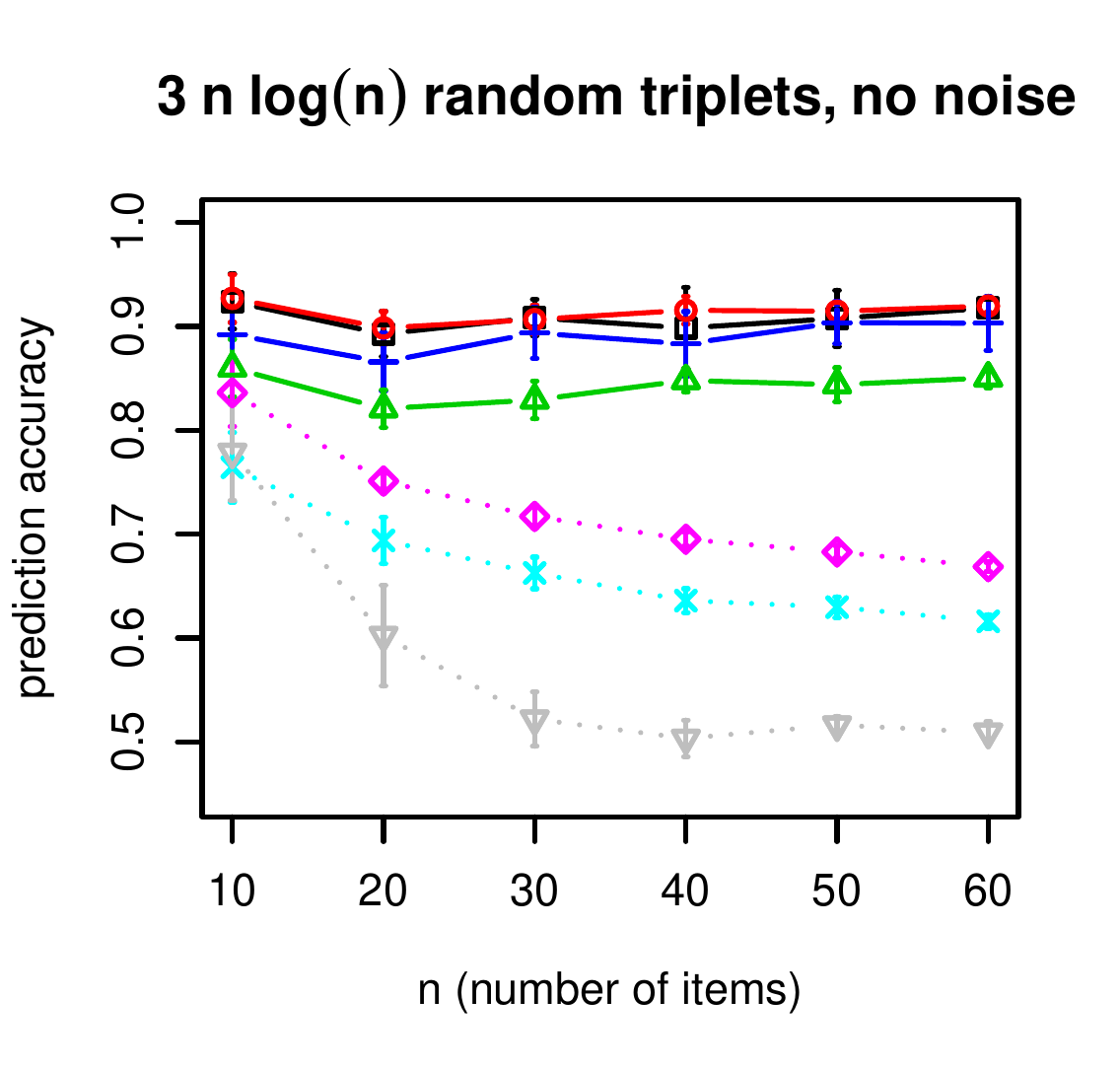}
    \label{fig:compare_3nlogn_synthetic_nonoise}
    \end{subfigure}
    \begin{subfigure}{0.32\textwidth}
    \centering
    \includegraphics[width=\linewidth]{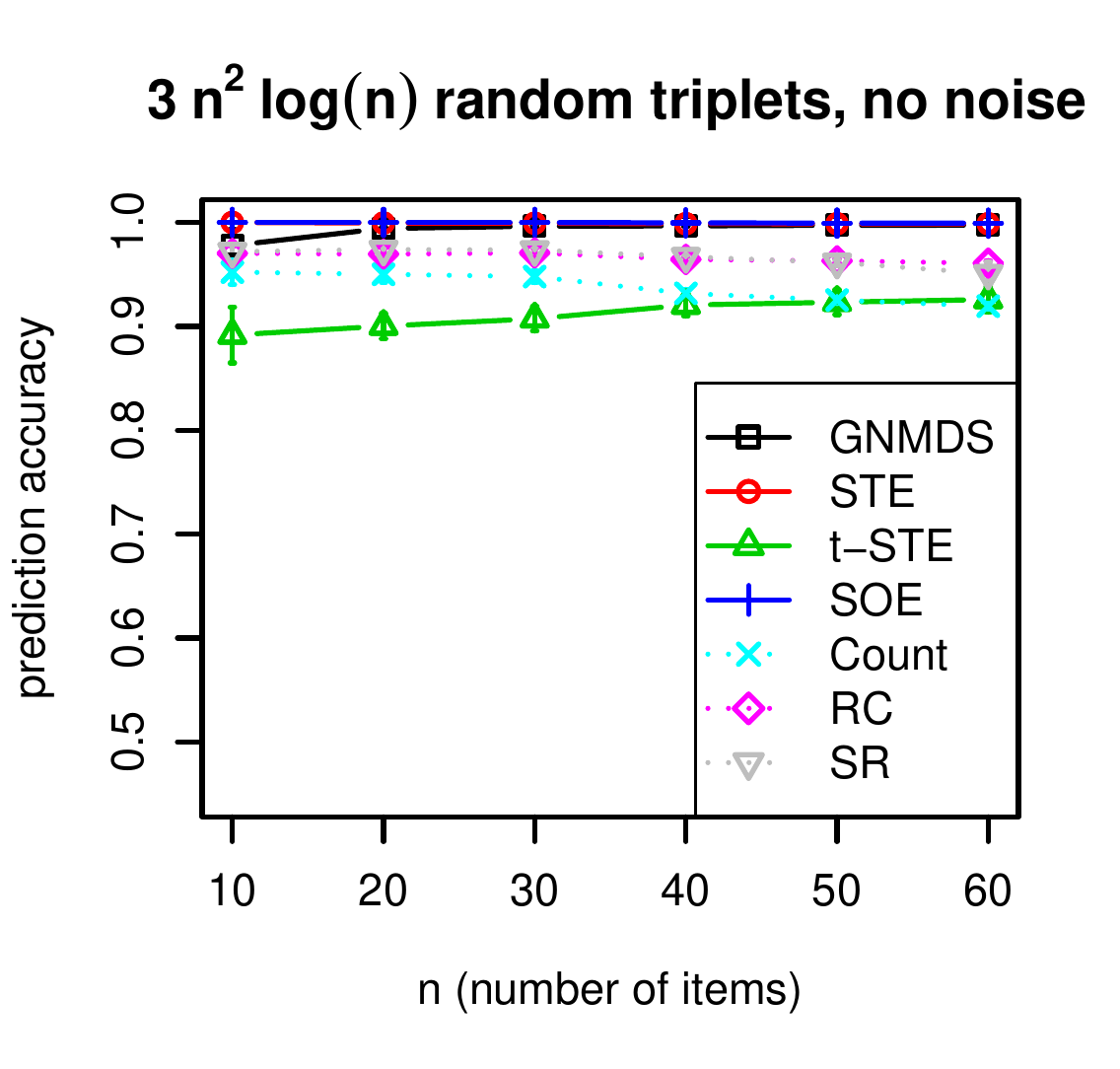}
    \label{fig:compare_3nlogn_synthetic_noise}
    \end{subfigure}
    \begin{subfigure}{0.32\textwidth}
    \centering
    \includegraphics[width=\linewidth]{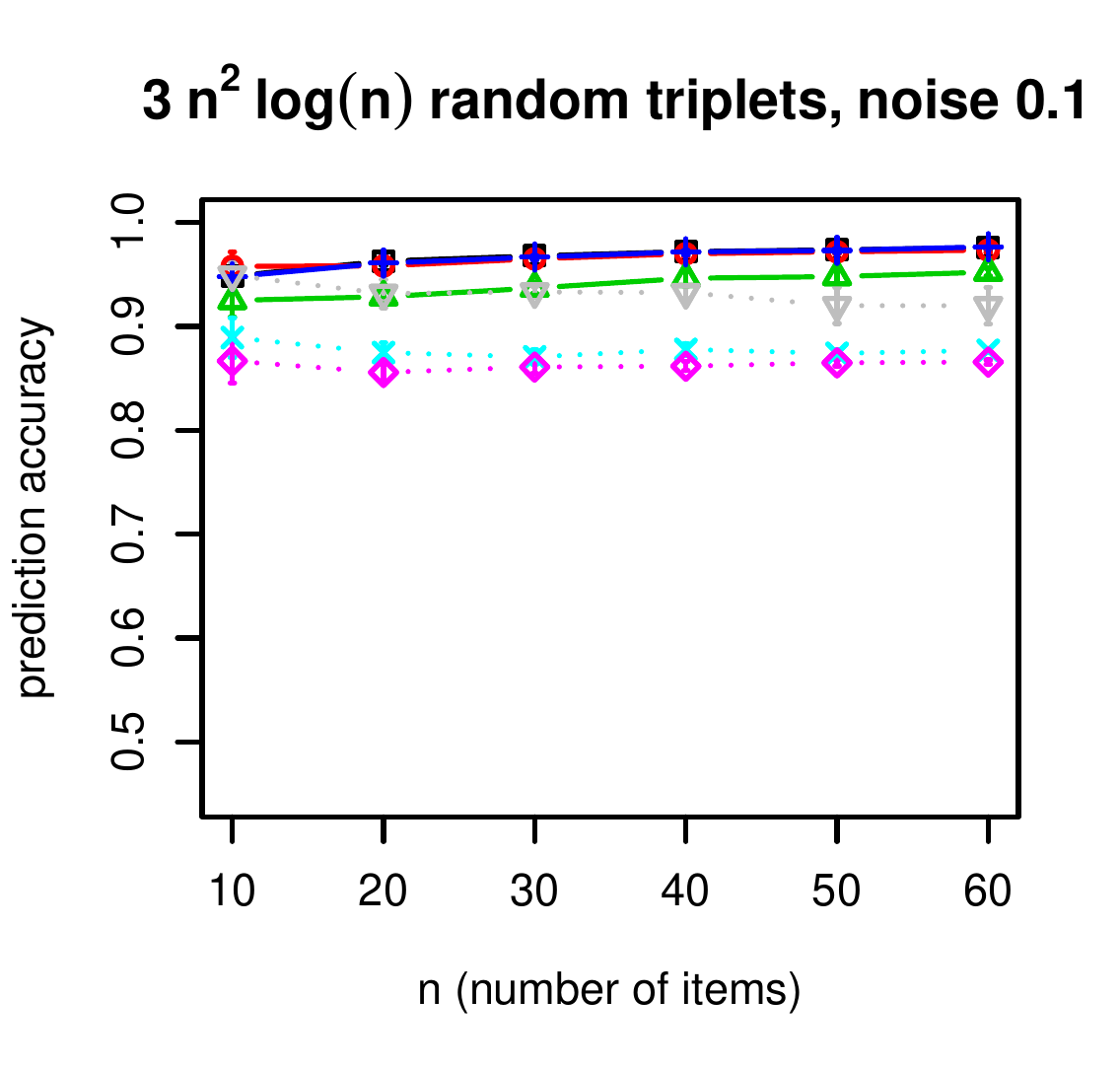}
    \label{fig:compare_3n2logn_synthetic_noise}
    \end{subfigure}
    
    \caption{Triplet prediction accuracy for various embedding and ranking methods as in \fig{fig:compare_methods_synthetic_eidolon_example}, but with different noise and different number of items (see figure titles). The number of triplets given to the algorithm depends on the number of items $n$, which varies along the horizontal axis. Items were generated as points uniformly at random in $[0,1]^3$ in each run of the simulation. We show the averages over 10 runs. Embedding methods embedded in 3 dimensions.  }
    \label{fig:compare_methods_synthetic_app}
\end{figure}

Here, we present complementary results of the simulations from in Section~\ref{sec:simlulations}. \fig{fig:compare_methods_synthetic_app} shows similar results as \fig{fig:compare_methods_synthetic_eidolon_example}, but in absence of noise or with $3n^2 \log_2 (n)$ known triplet comparisons instead of just $3n \log_2 (n)$. We see that with $3n^2 \log_2 (n)$ triplets, ranking methods perform well and can be useful if no geometrical structure that could be exploited by embedding approaches is present in the data. However, collecting $3n^2 \log_2 (n)$ triplets may be infeasible for tasks that involve even just a moderately large number of items.
 
\begin{figure}
    \centering
    \begin{subfigure}{0.49\textwidth}
    \centering
    \includegraphics[width=\linewidth]{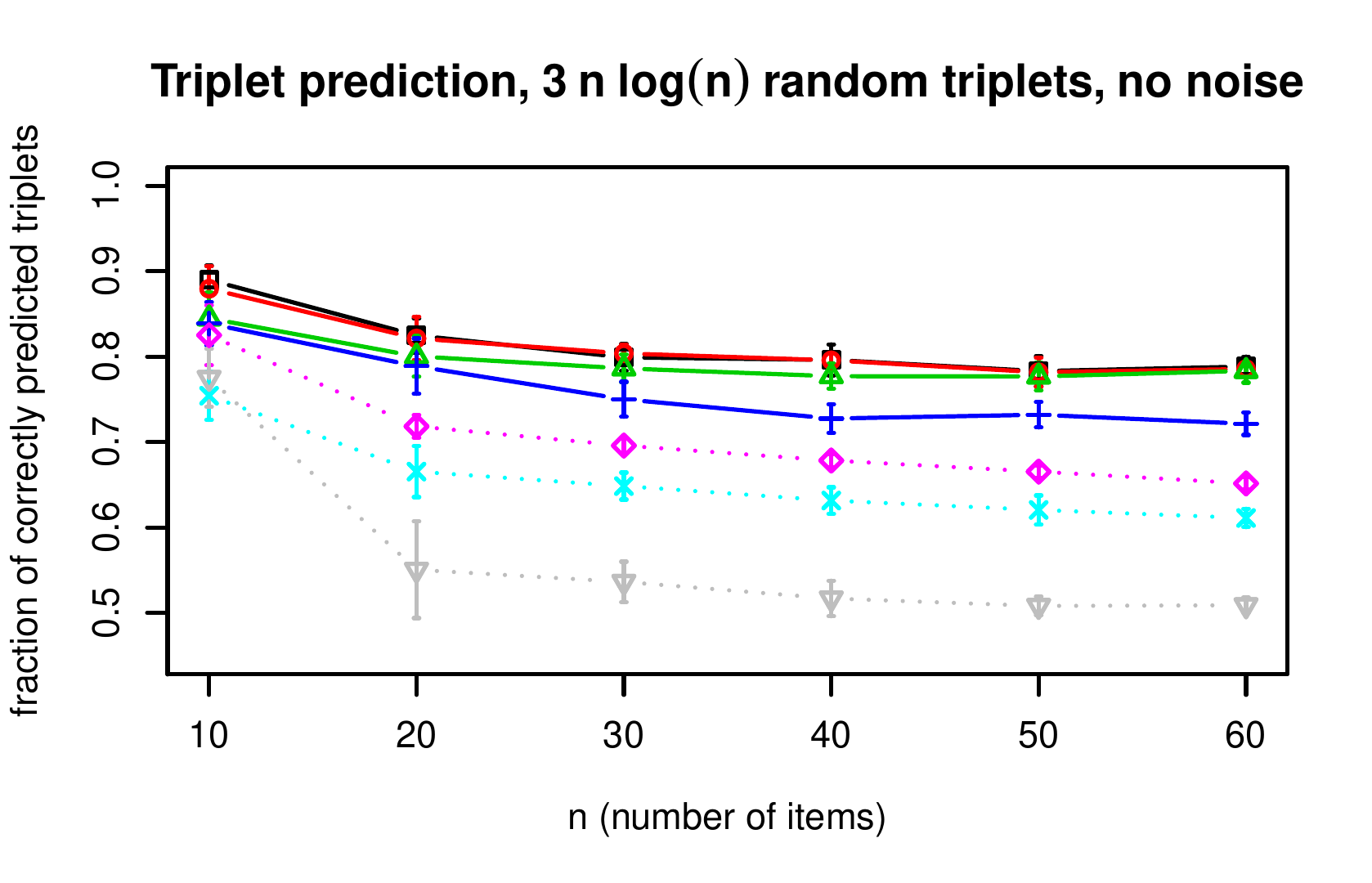}
    \label{fig:compare_3nlogn_mnist_nonoise}
    \end{subfigure}
    \begin{subfigure}{0.49\textwidth}
    \centering
    \includegraphics[width=\linewidth]{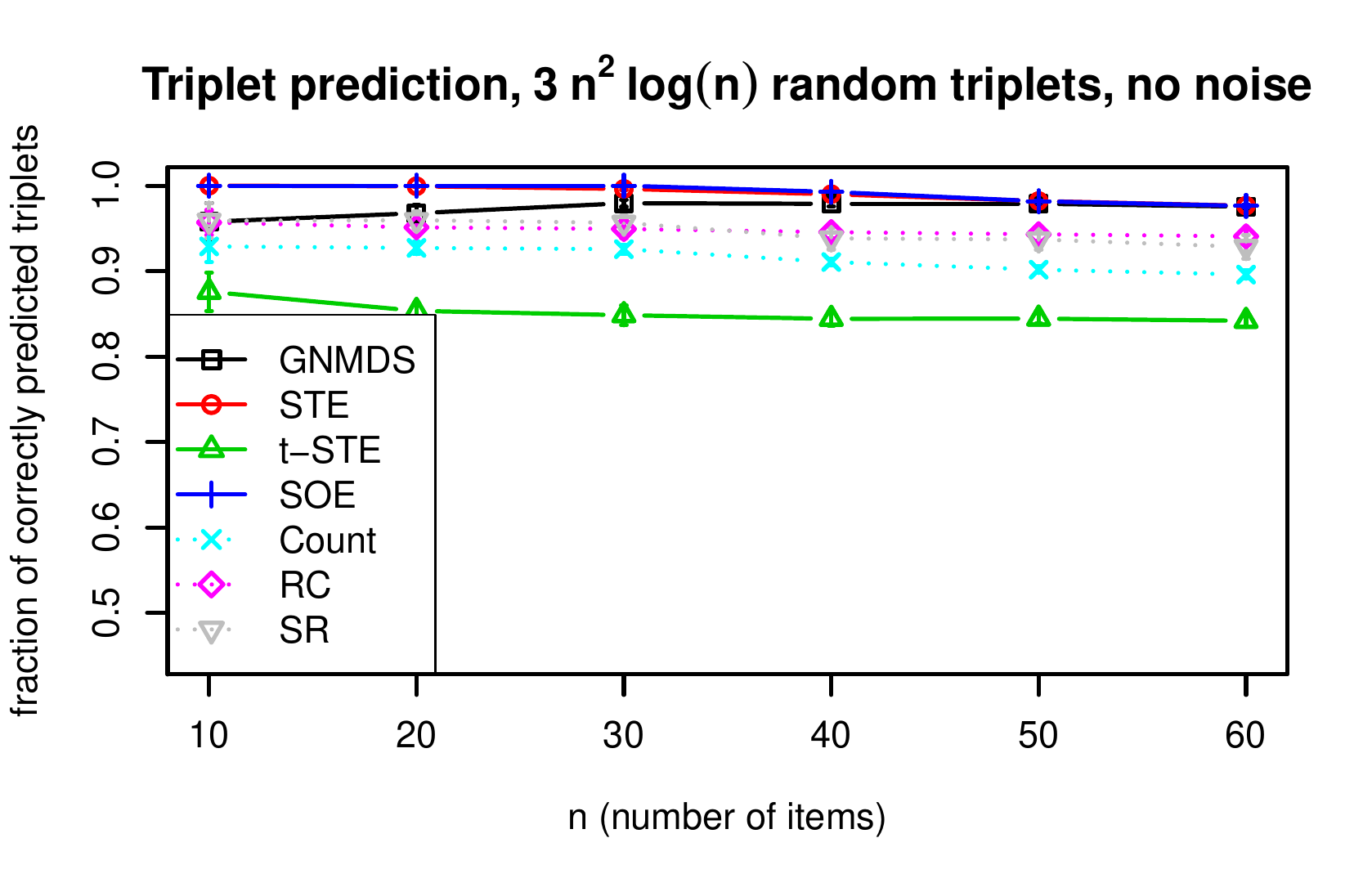}
    \label{fig:compare_3n2logn_mnist_nonoise}
    \end{subfigure}
    \begin{subfigure}{0.49\textwidth}
    \centering
    \includegraphics[width=\linewidth]{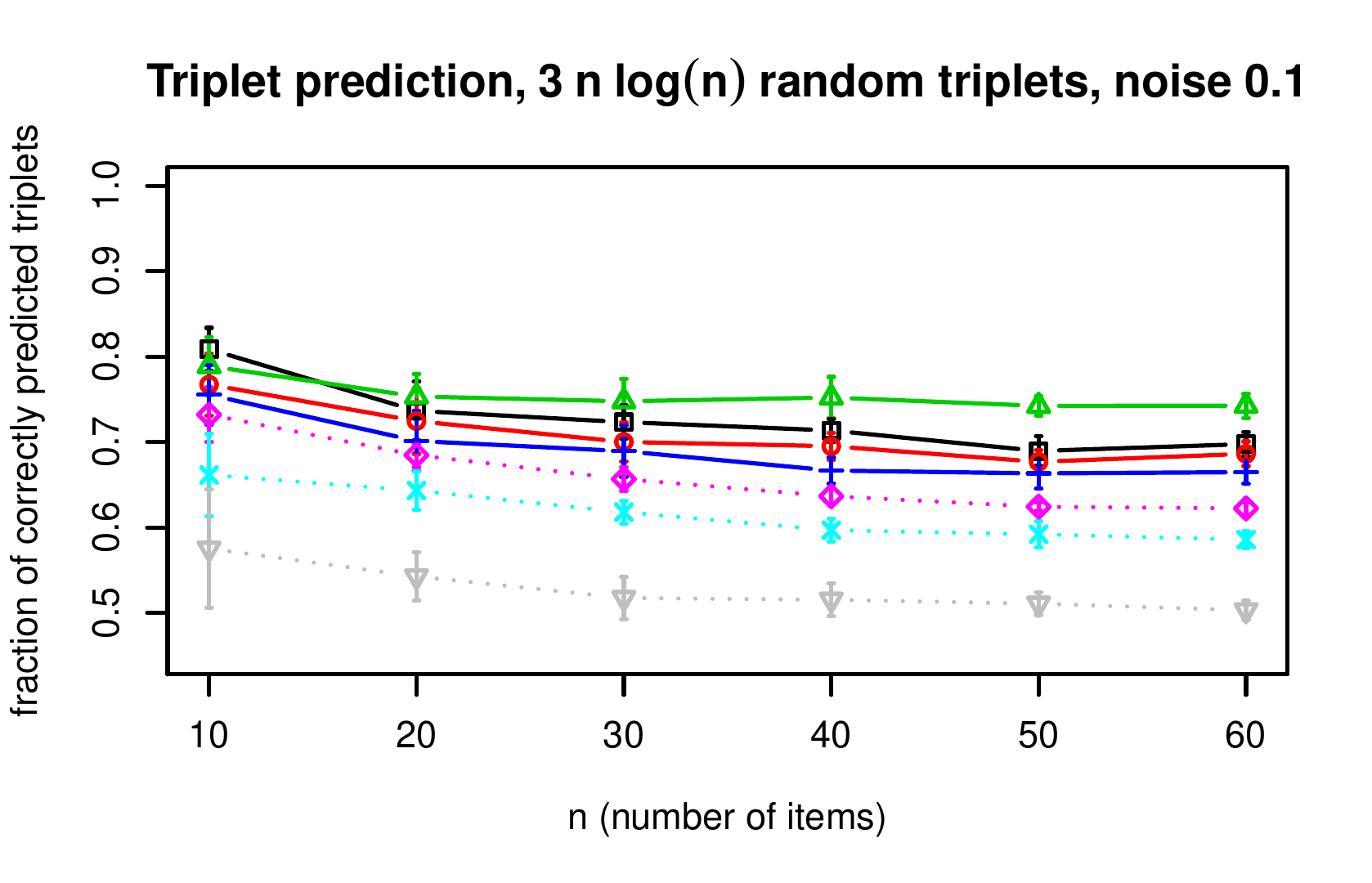}
    \label{fig:compare_3nlogn_mnist_noise}
    \end{subfigure}
    \begin{subfigure}{0.49\textwidth}
    \centering
    \includegraphics[width=\linewidth]{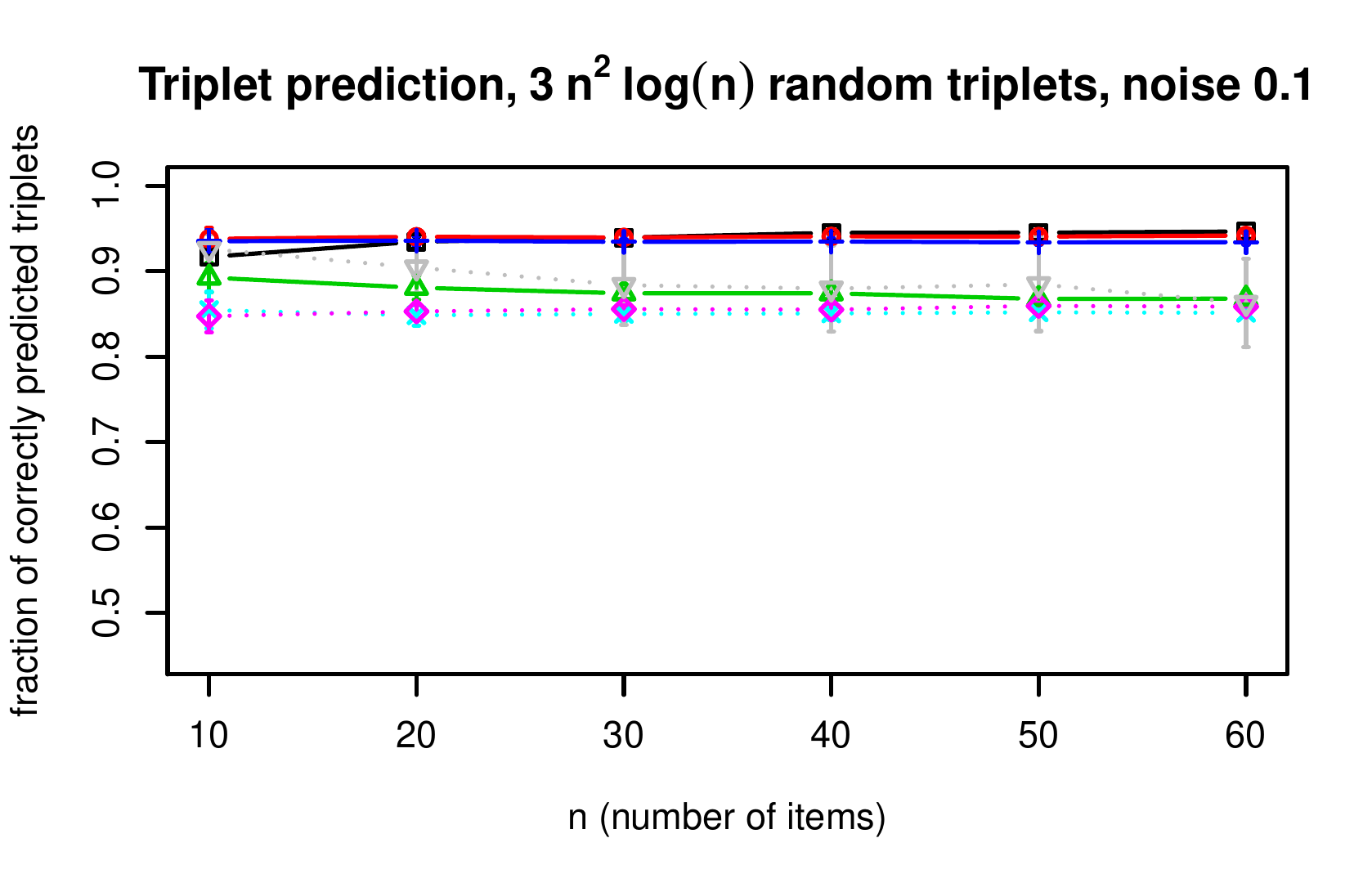}
    \label{fig:compare_3n2logn_mnist_noise}
    \end{subfigure}
    
    \caption{Triplet prediction accuracies as in \fig{fig:compare_methods_synthetic_eidolon_example} (left) and \fig{fig:compare_methods_synthetic_app}, but with items being MNIST digits (chosen uniformly at random from all 70000 MNIST digits for each run). Embedding methods embedded in 20 dimensions.}
    \label{fig:compare_methods_mnist}
\end{figure}

In \fig{fig:compare_methods_mnist}, we run the same simulation on MNIST data instead of synthetically generated data points. We still use Euclidean distance as ground truth, but embed in 20 dimensions to account for the higher dimensionality of the MNIST data (the raw data is 784-dimensional). The results are similar to those with synthetically generated data, showing that the apporaches we present in this work are able to produce meaningful results for a large variety of data. Due to the higher dimensionality of the data, the prediction accuracy is slightly worse as the number of training triplets is the same as in the above simulation.

\begin{figure}
    \centering
    \begin{subfigure}{0.49\textwidth}
    \centering
    \includegraphics[width=\linewidth]{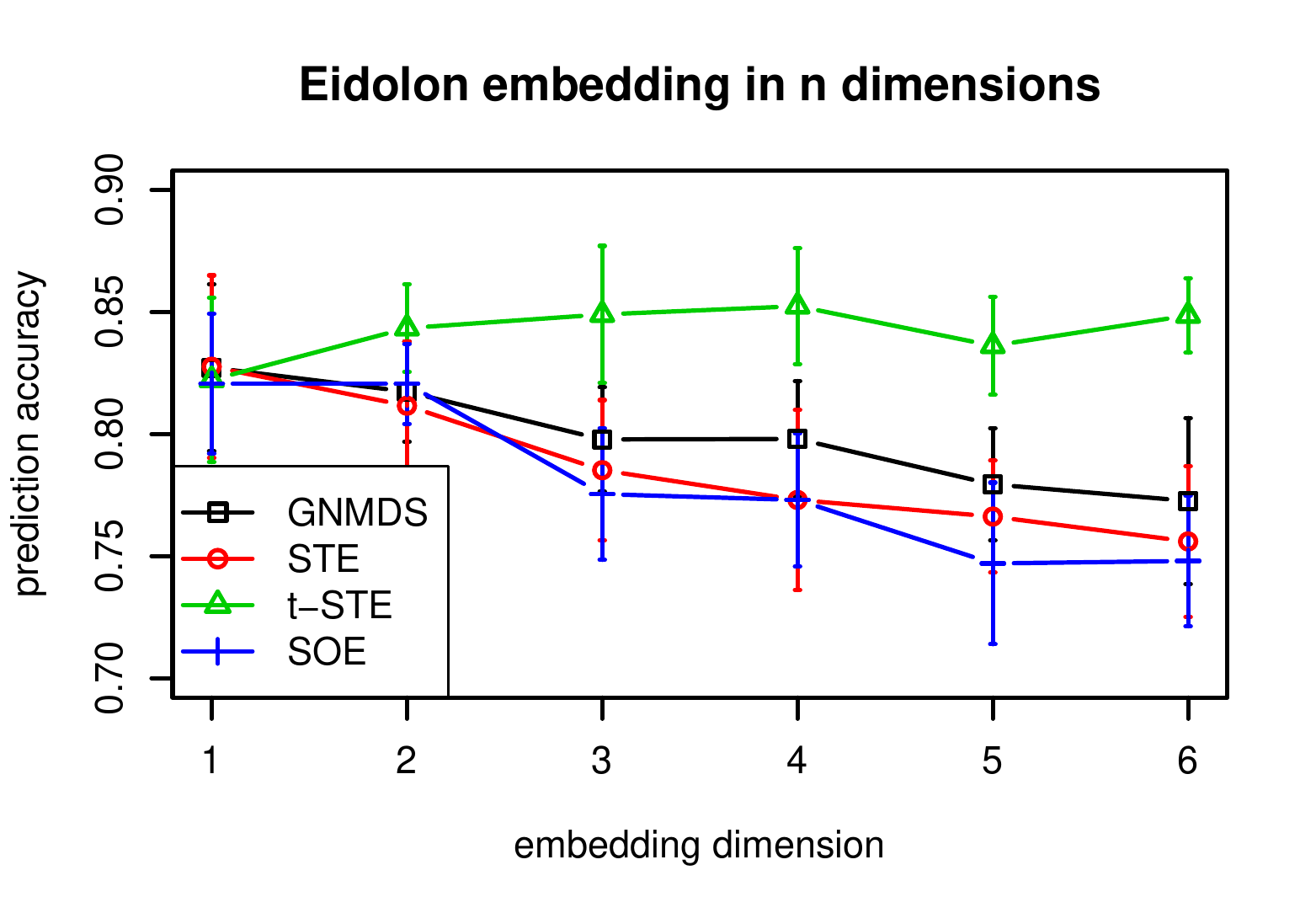}

    \end{subfigure}
    \begin{subfigure}{0.49\textwidth}
    \centering
    \includegraphics[width=\linewidth]{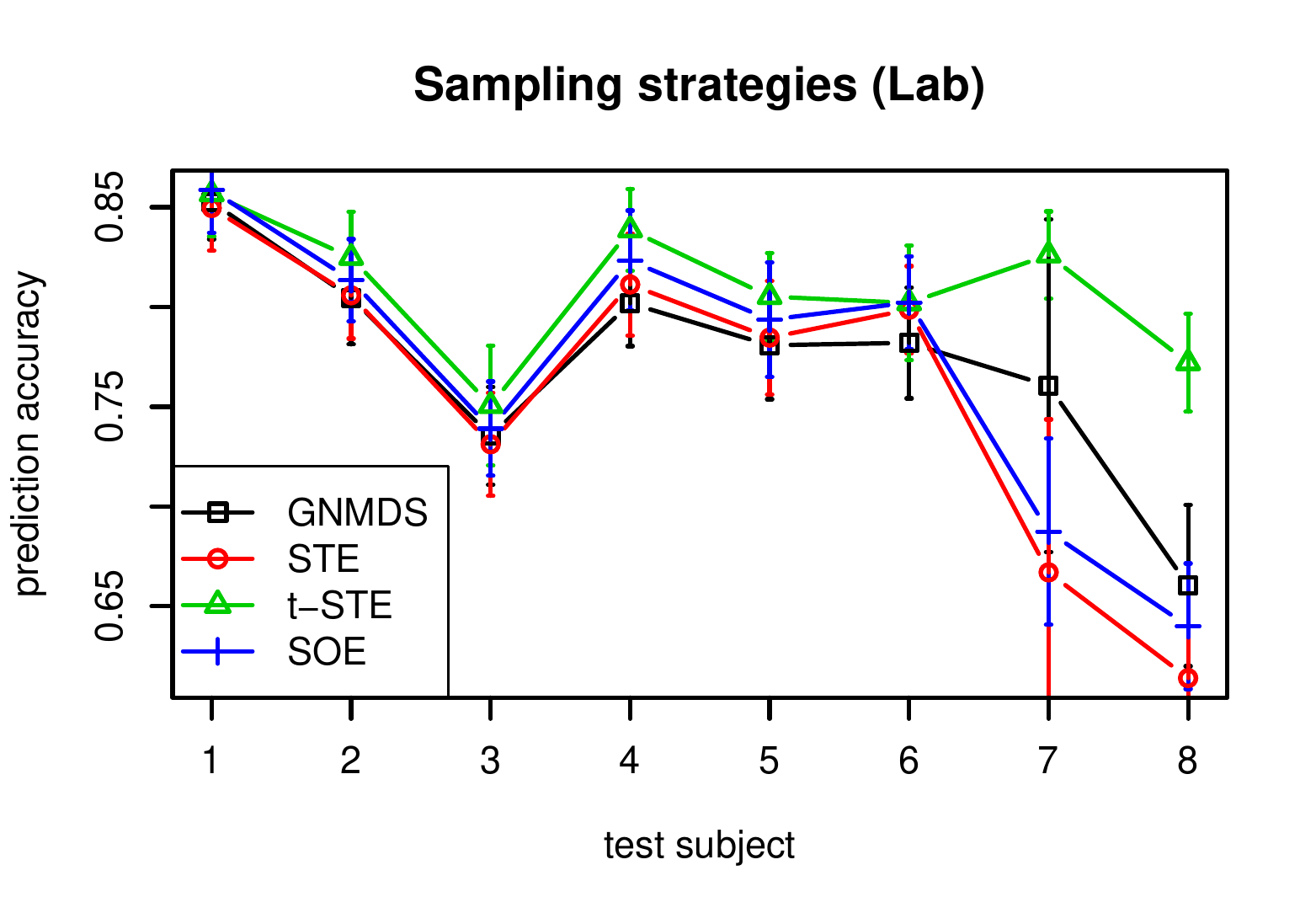}
    \end{subfigure}
    
    \caption{Left: We use embedding in different dimensions for triplet prediction. 1500 triplets from lab subject 1 are used for training, 250 for testing (as in several of our evaluations). Right: We check how well 4500 triplets of a single subject from the lab experiment can predict the a test set of 250 triplets of the same subject.}
    \label{fig:embedding_dimensions_lab}
\end{figure}

\subsection{Lab / MTurk experiments}
\label{app:lab_mturk}

\begin{figure}
\centering
\includegraphics[width=1\linewidth]{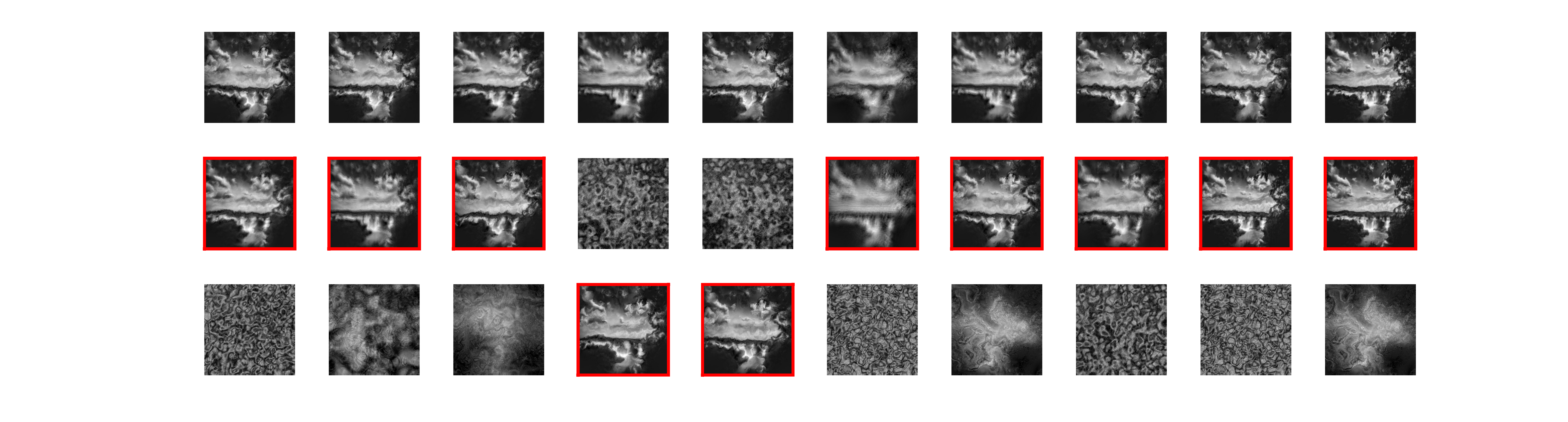}
\caption{10 out of 20 triplet questions used as Gold Standards in one of the MTurk experiments. Each column corresponds to one triplet question. The first row is the query image. The second and third row show the options of the participant. The obvious answers are denoted by red frames}
\label{fig:ex_gold_standards}
\end{figure}

A set of 10 obvious Gold Standard questions is shown in \fig{fig:ex_gold_standards}. Each column represents a triplet question. The first row is the query image. The participant should choose one of the images in the next two rows as a response. The obvious response the each query is denoted by a red frame. These triplets were randomly distributed in the sessions and they were asked with the same layout.

\fig{fig:embedding_dimensions_lab} (left) shows the prediction result of embedding for subject 1 of the lab experiment with 1500 triplets as training set and 250 triplets as test set. Increasing the dimension from 1 to 2 improves the prediction accuracy significantly, however more than 2 dimensions cannot lead to a considerable improvement. We also observe that t-STE outperforms the other methods in all dimensions.

\fig{fig:embedding_dimensions_lab} (right) shows the detailed prediction accuracy of various subsampling strategies when we use 4500 triplets as training and 250 triplets as test. This figure is the expansion of the box plots presented in \fig{fig:repeated_lab_mturk} (b).  

\begin{figure}
    \centering
    \begin{subfigure}{0.49\textwidth}
    \centering
    \includegraphics[height=4.3cm]{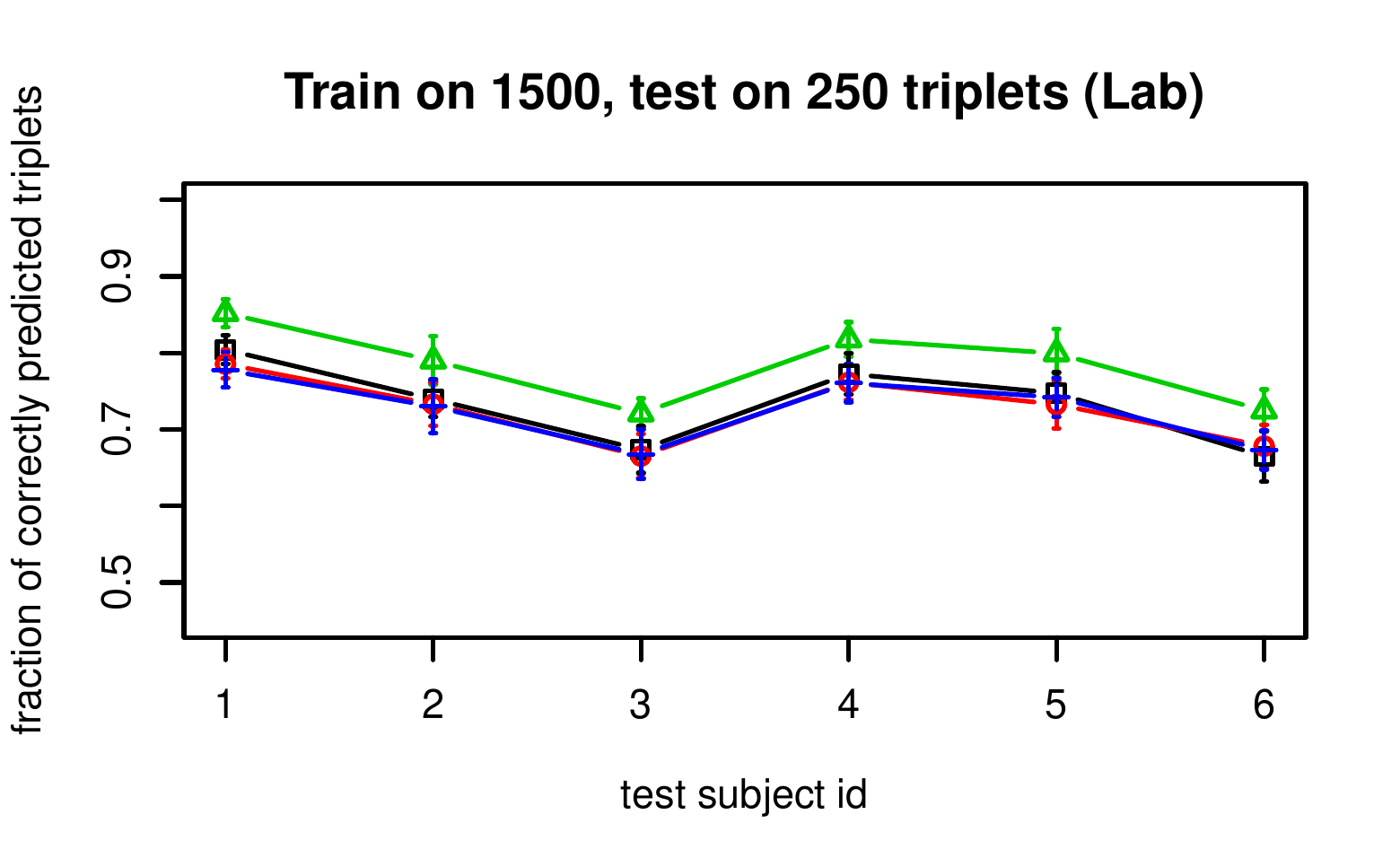}
    \end{subfigure}
    \begin{subfigure}{0.49\textwidth}
    \centering
    \includegraphics[height=4.3cm]{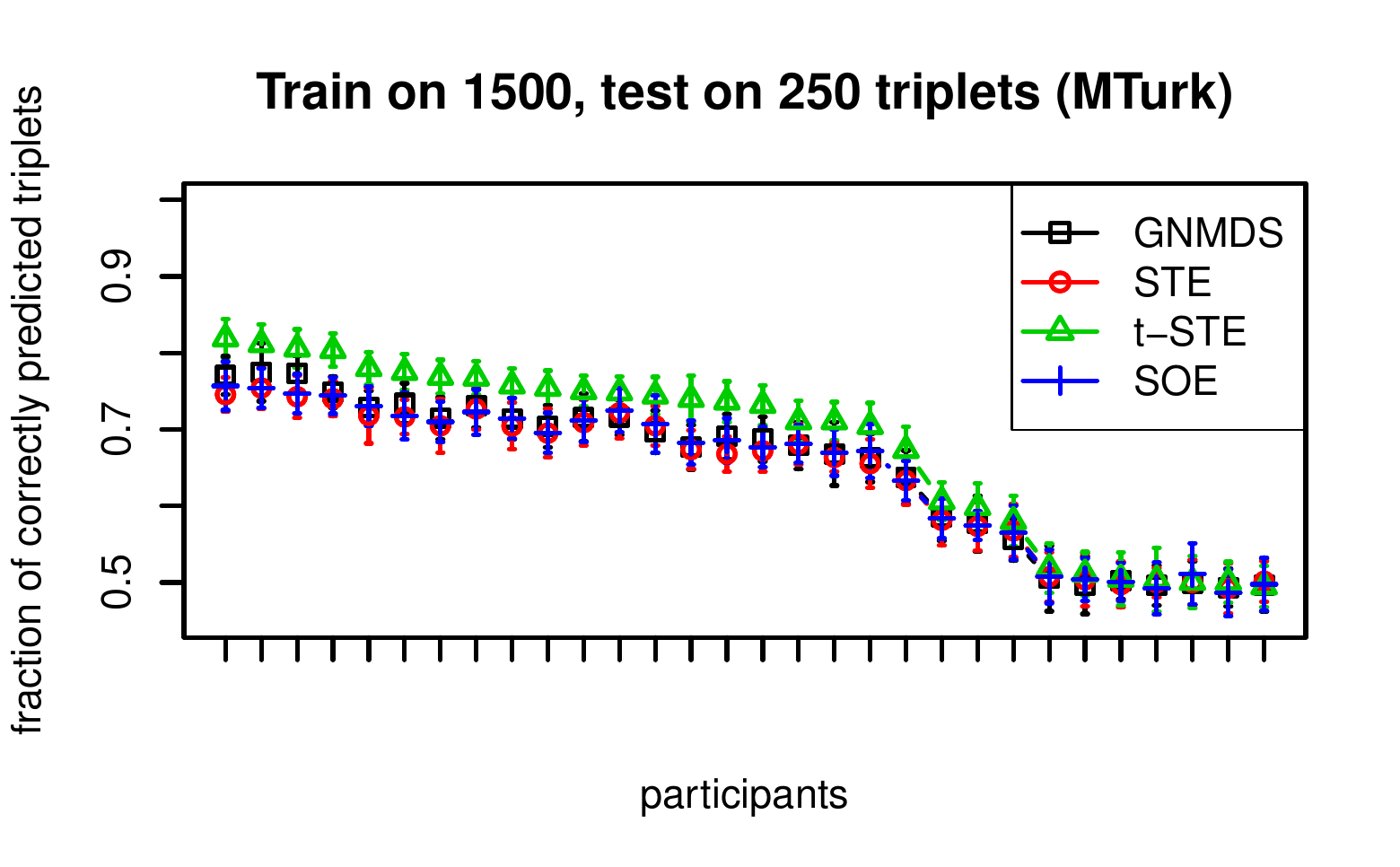}
    \end{subfigure}
    
    \caption{We check how well the triplets of a single subject can predict the triplets of the same subject. From the set of 2000 triplets of a subject (we only use the first 2000 for lab experiment participants), we select 250 triplets at random as validation set, select 1500 triplets from the rest as training set for embedding and report the accuracy on the validation set. The accuracies are averaged over 20 repetitions. Left: Lab participants 1--6; Right: MTurk participants, sorted by t-STE triplet prediction accuracy.}
    \label{fig:accuracy_lab_mturk_single}
\end{figure}

\fig{fig:accuracy_lab_mturk_single} demonstrates the detailed prediction accuracy of random subsampling for both lab and MTurk experiments. Each point corresponds to one session of 2000 triplets. The MTurk participants are sorted based on the performance of t-STE method on the horizontal axis. The last 7 subjects are exactly the same subject who did not pass the sanity check. Most of the other MTurk participants have acceptable accuracy. This figure is the expansion of the two left box plots presented in \fig{fig:repeated_lab_mturk} (a).

\begin{figure}
\centering
    \begin{subfigure}{0.32\textwidth}
    \centering
    \includegraphics[width=\linewidth]{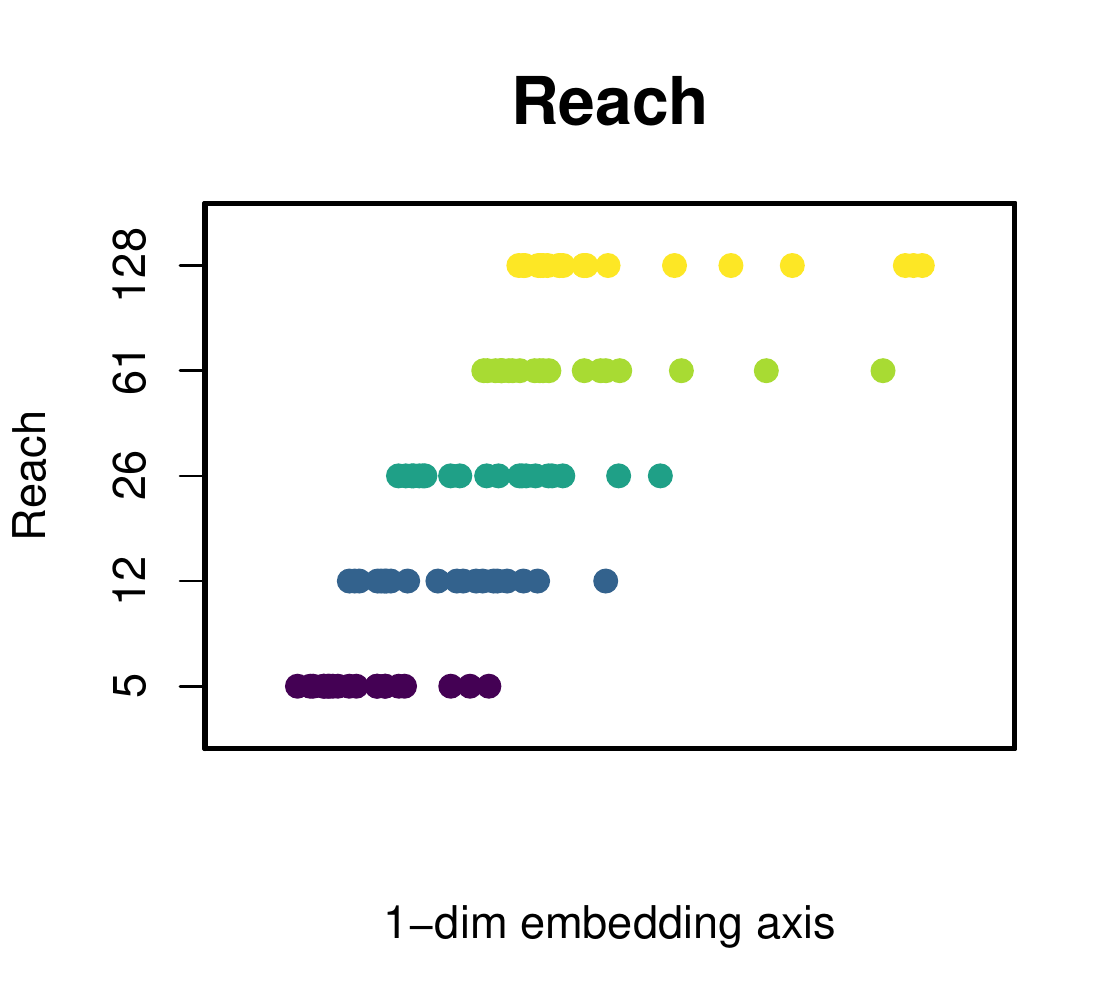}
    \end{subfigure}
    \centering
    \begin{subfigure}{0.32\textwidth}
    \centering
    \includegraphics[width=\linewidth]{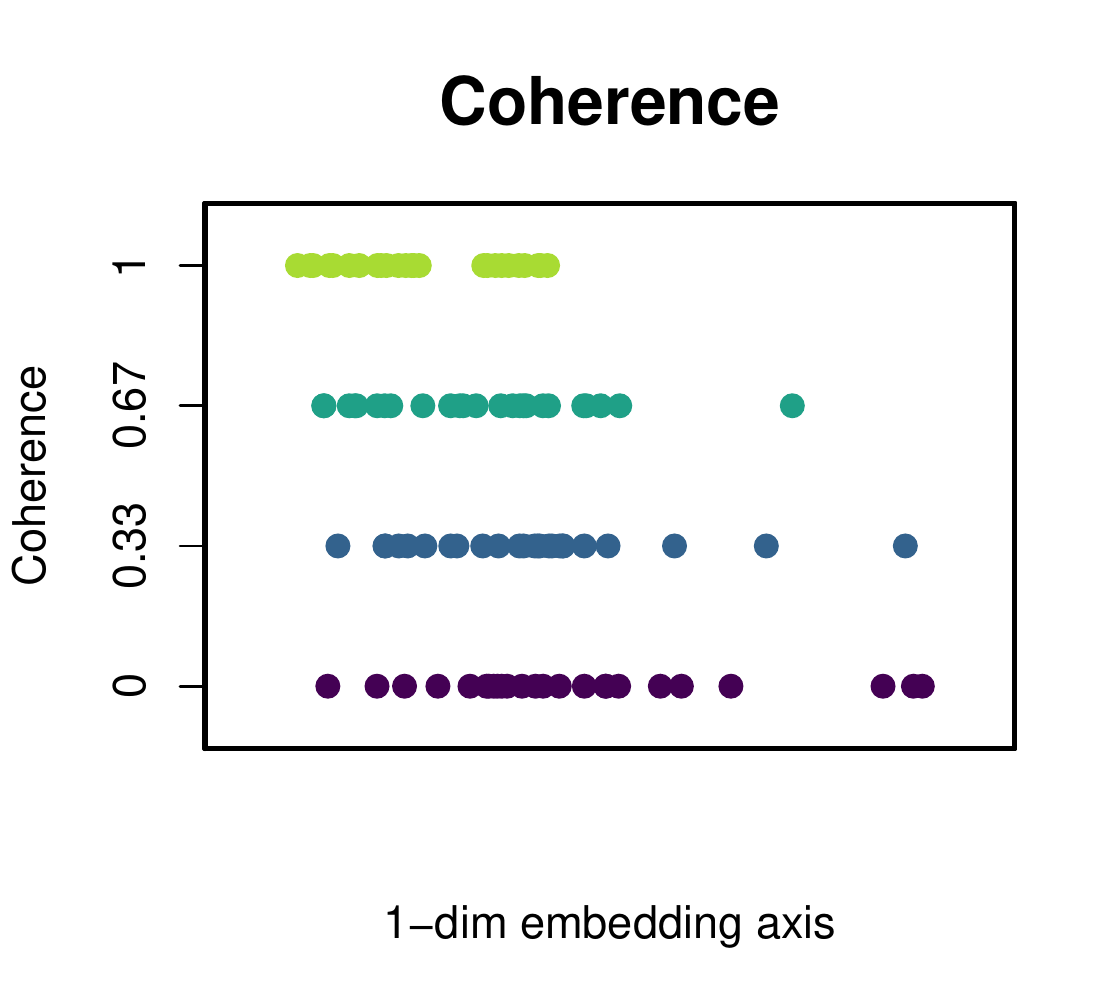}
    \end{subfigure}
    \begin{subfigure}{0.32\textwidth}
    \centering
    \includegraphics[width=\linewidth]{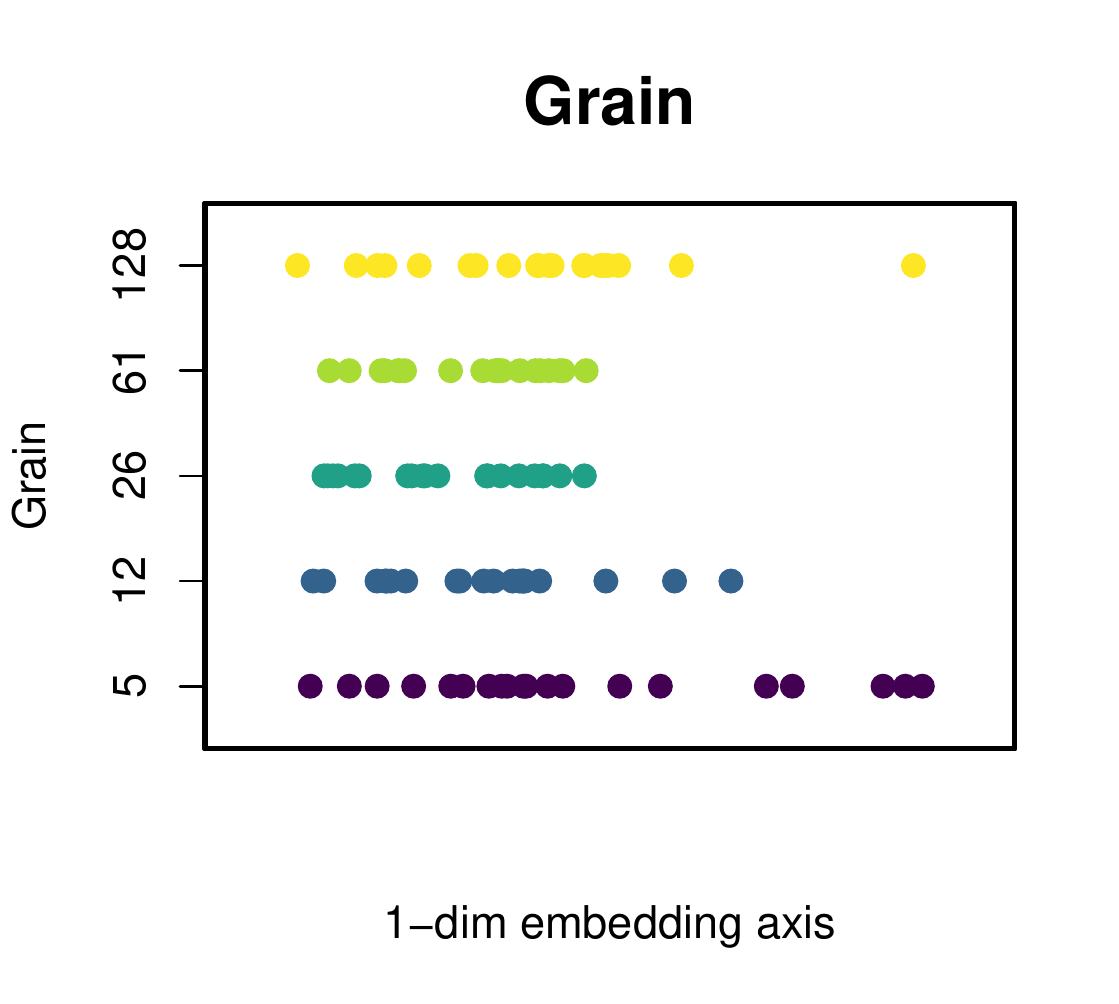}
    \end{subfigure}
    \centering
    \caption{t-STE embedding in $1$ dimensions from all triplets by the first three lab subjects (i.e., all sessions with completely random triplets). Each dot corresponds to one of the 100 images from our eidolon experiment, dots are colored by the values of the three parameters reach, coherence and grain used to generate the images. The vertical coordinate is also determined by the parameter value.}
    \label{fig:parameters_in_embedding_lab_1dim}
\end{figure}

\begin{figure}
\centering
    \begin{subfigure}{0.32\textwidth}
    \centering
    \includegraphics[width=\linewidth]{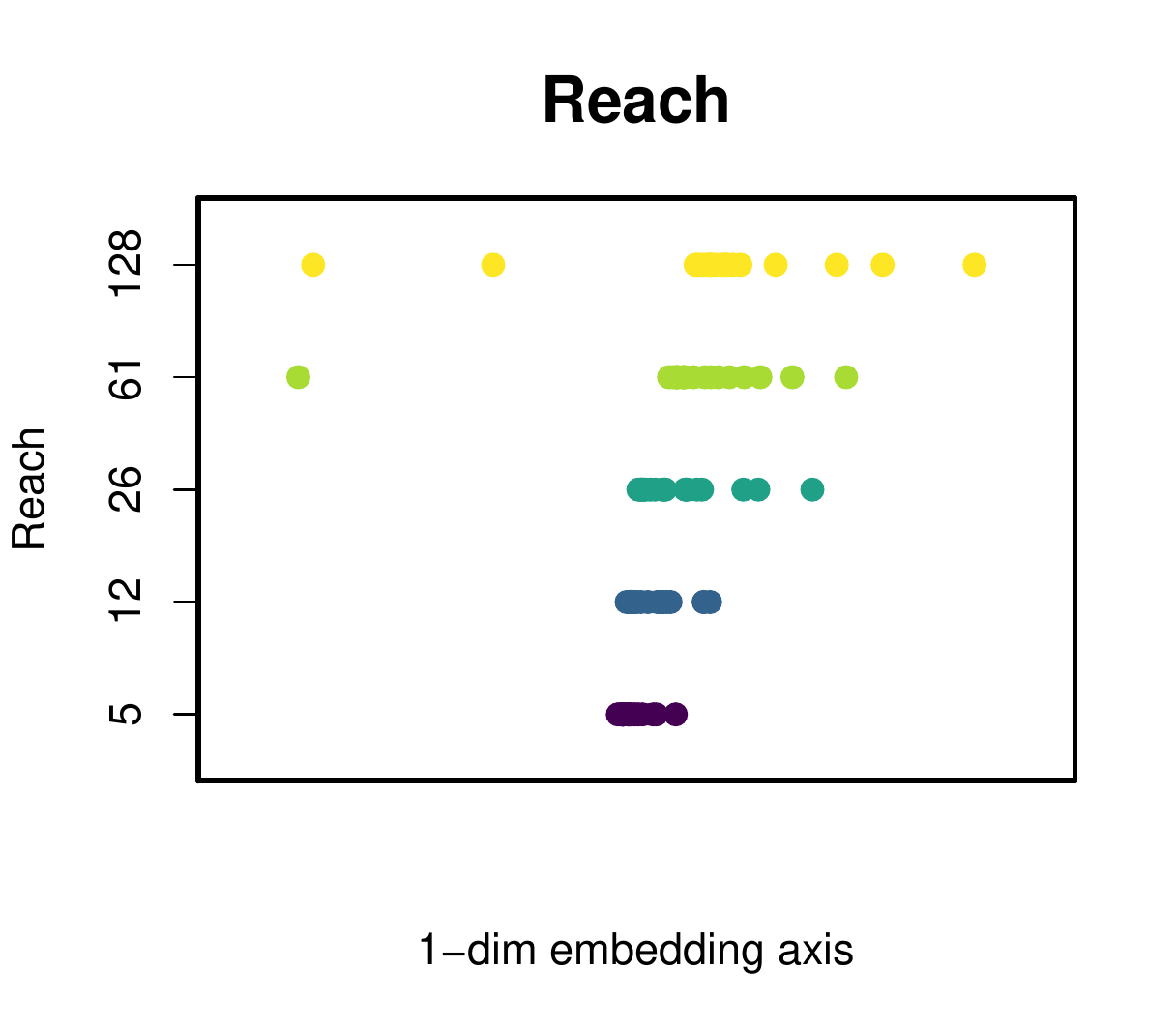}
    \end{subfigure}
    \centering
    \begin{subfigure}{0.32\textwidth}
    \centering
    \includegraphics[width=\linewidth]{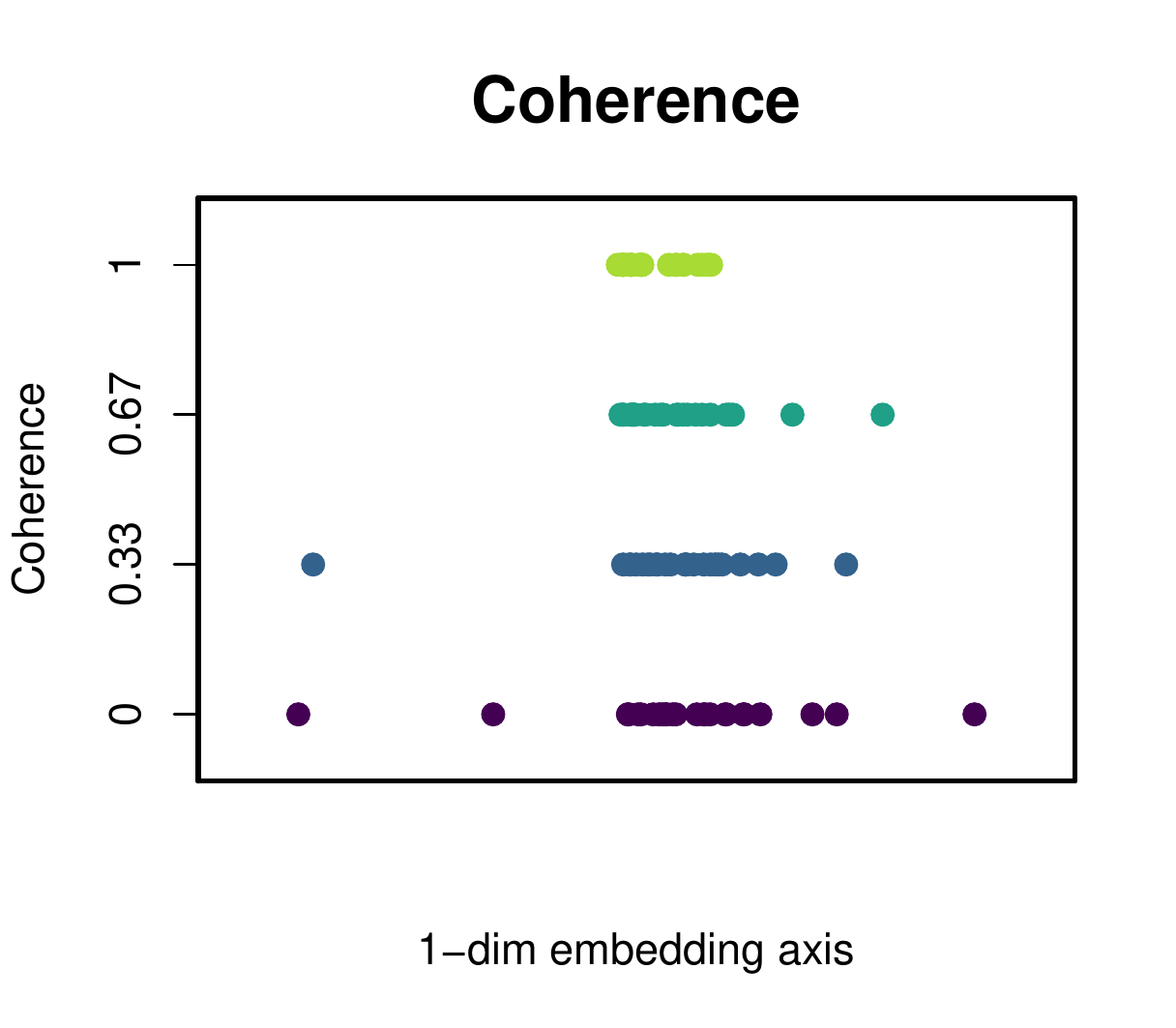}
    \end{subfigure}
    \begin{subfigure}{0.32\textwidth}
    \centering
    \includegraphics[width=\linewidth]{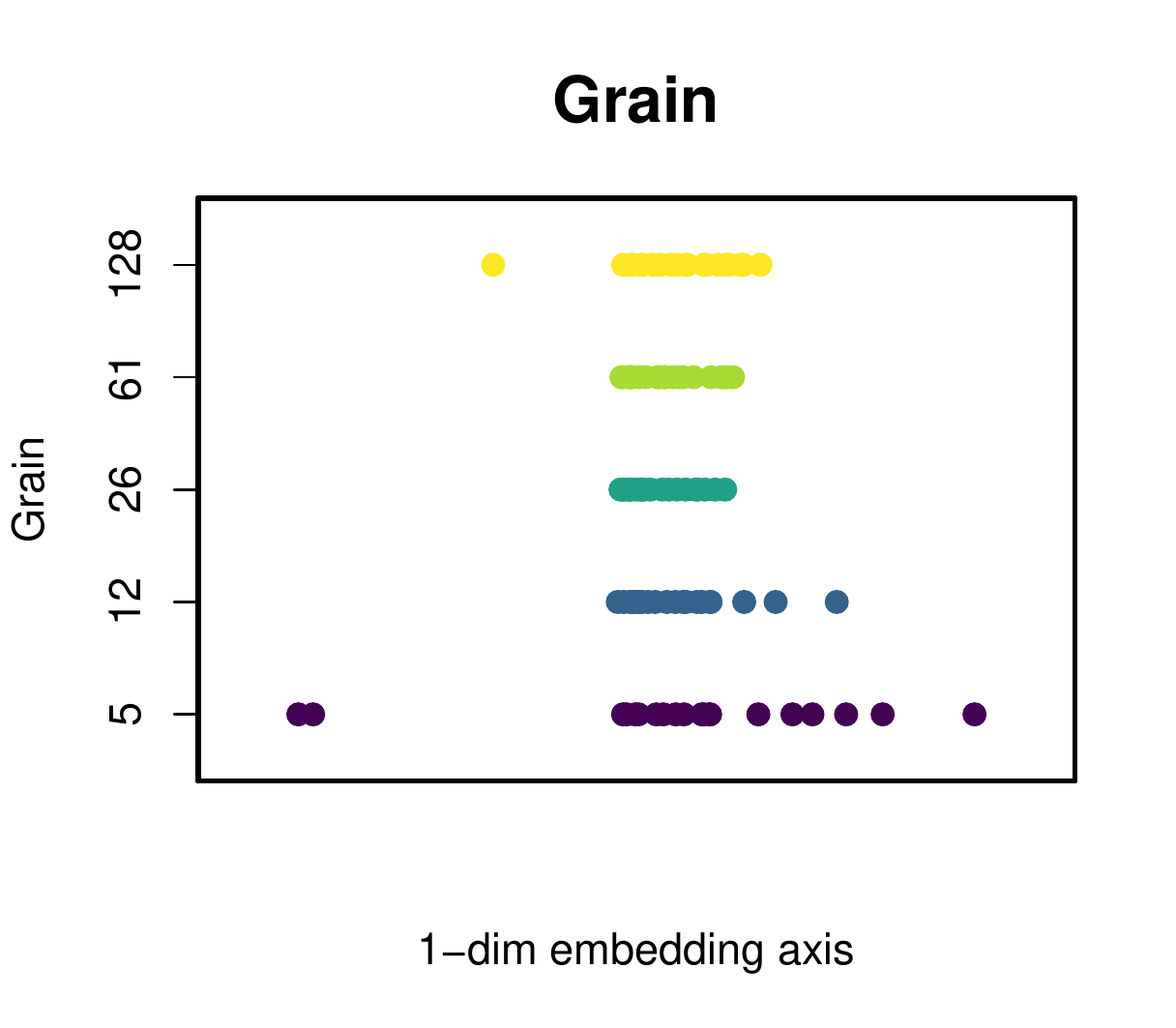}
    \end{subfigure}
    \centering
    \caption{t-STE embedding in $1$ dimension from the triplets of all 23 MTurk participants from the random setting who passed the sanity check. Each dot corresponds to one of the 100 images from our eidolon experiment, dots are colored by the values of the three parameters reach, coherence and grain used to generate the images. The vertical coordinate is also determined by the parameter value.}
    \label{fig:parameters_in_embedding_mturk_1dim}
\end{figure}

\fig{fig:parameters_in_embedding_lab_1dim} shows the embedding in one dimension based on the triplets of the first three subjects in the lab who answered the random sabsampled triplets. We denoted the same embedding for the 23 MTurk participants (who passed the sanity check) in the random subsampling experiment in \fig{fig:parameters_in_embedding_mturk_1dim}. The lab and MTurk embeddings look quite similar, except that we have more outliers in MTurk experiments. Similar to the two dimensional embedding, the reach parameter seems to have the most correlation with the embedding.

\begin{figure}
\centering
    \begin{subfigure}{0.32\textwidth}
    \centering
    \includegraphics[width=\linewidth]{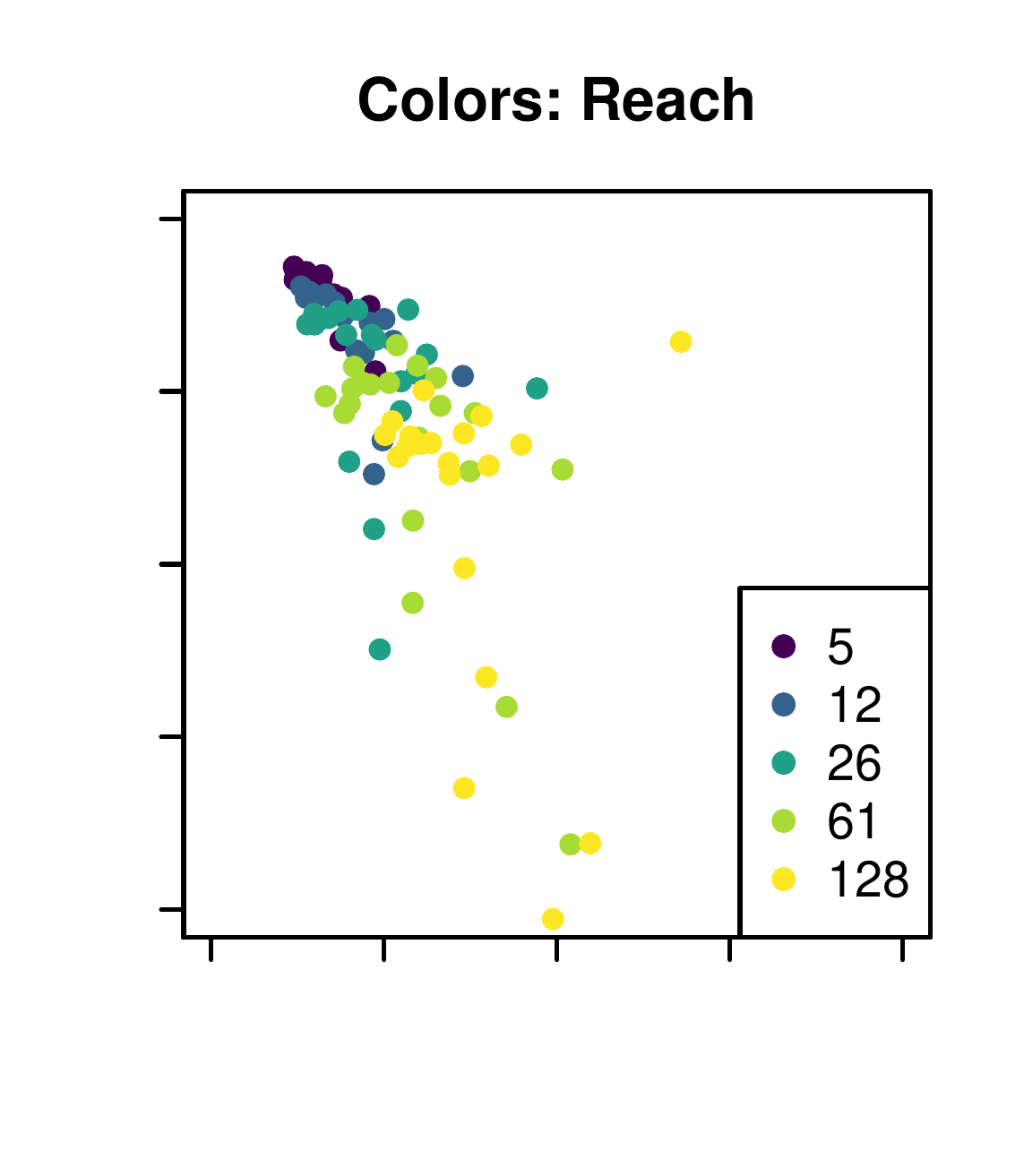}
    \end{subfigure}
    \centering
    \begin{subfigure}{0.32\textwidth}
    \centering
    \includegraphics[width=\linewidth]{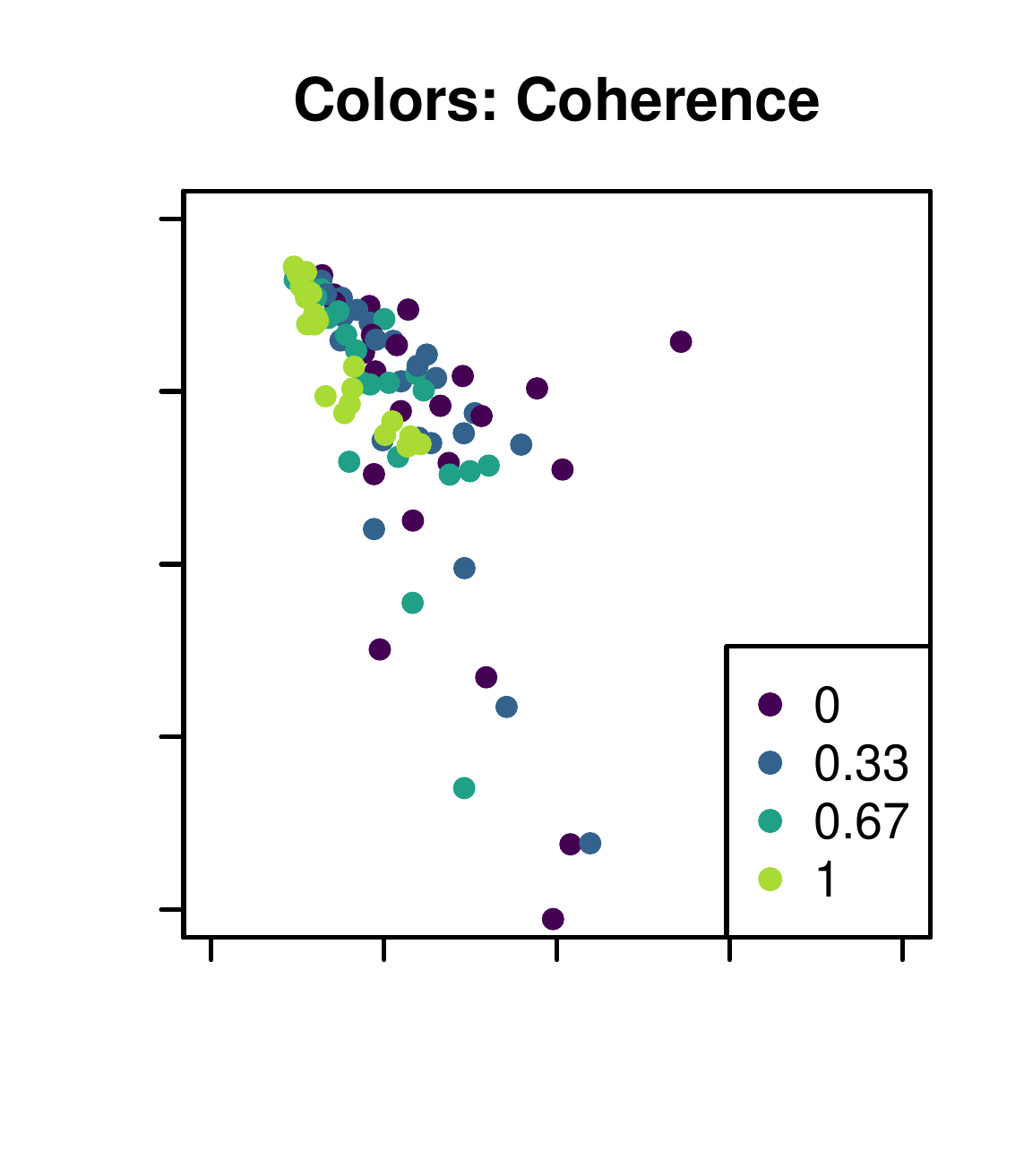}
    \end{subfigure}
    \begin{subfigure}{0.32\textwidth}
    \centering
    \includegraphics[width=\linewidth]{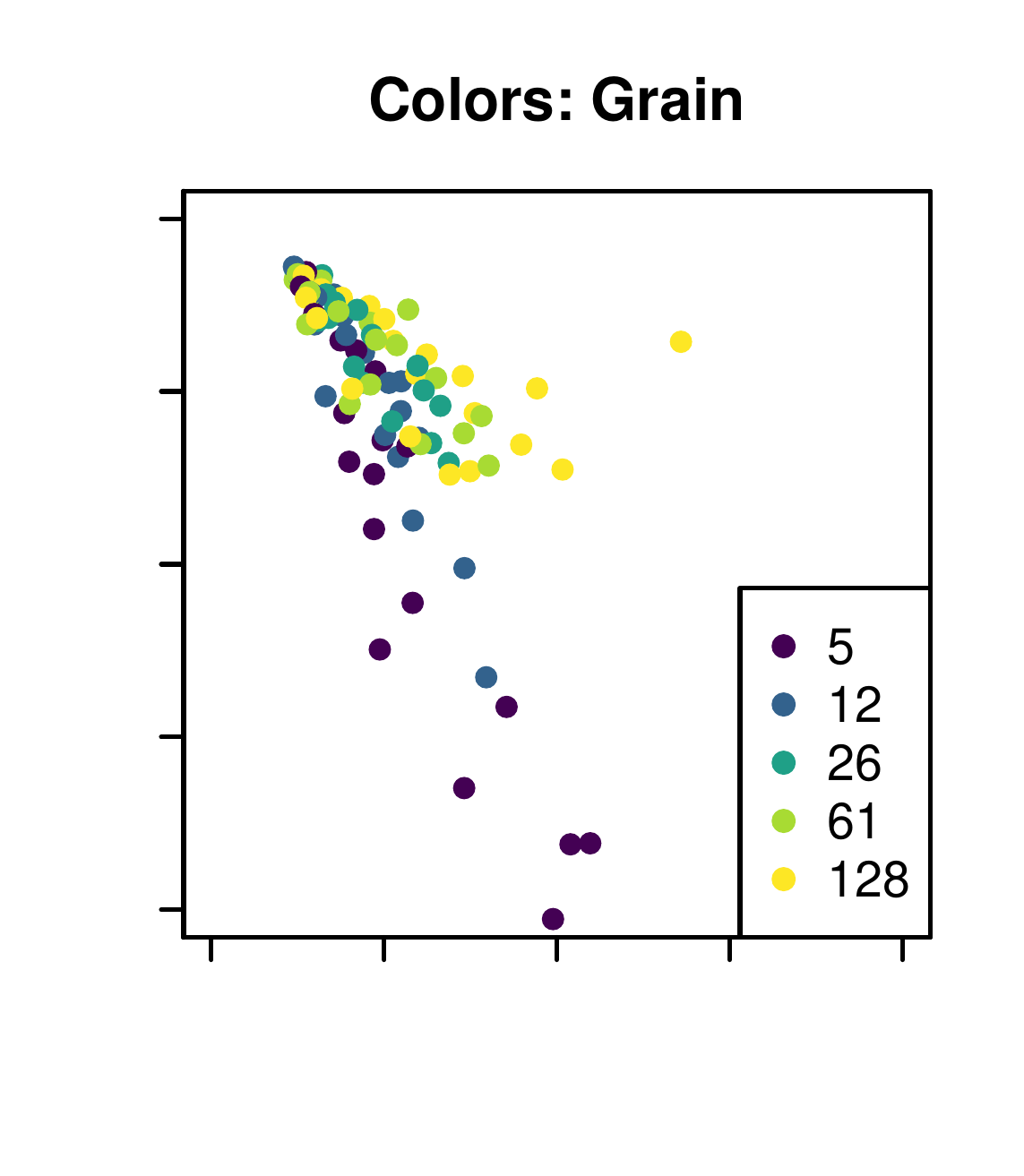}
    \end{subfigure}
    \centering
    \caption{t-STE embedding in $2$ dimensions from the triplets of all 23 MTurk participants from the random setting who passed the sanity check. Each dot corresponds to one of the 100 images from our eidolon experiment, dots are colored by the values of the three parameters reach, coherence and grain used to generate the images.}
    \label{fig:parameters_in_embedding_mturk}
\end{figure}

The two dimensional embedding in \fig{fig:parameters_in_embedding_mturk} is based on the randomly subsampled triplets of 23 MTurk participants who passed the sanity check. The results are very similar to the embedding of lab experiments presented in \fig{fig:parameters_in_embedding} disregarding the rotation and scaling. Note that ordinal information do not capture scaling and rotation information in Euclidean space.

\end{document}